\documentclass[10pt,journal,compsoc]{IEEEtran}

\ifCLASSOPTIONcompsoc
  \usepackage[nocompress]{cite}
\else
  \usepackage{cite}
\fi

\usepackage{amsmath,amsfonts,amsthm}
\usepackage{amssymb}
\newtheorem{theorem}{Theorem}

\usepackage{algorithm}
\usepackage[noend]{algpseudocode}
\usepackage{xcolor}
\usepackage{graphicx}
\usepackage{subfig}
\usepackage{booktabs}

\DeclareMathOperator*{\argmax}{\arg\!\max}
\DeclareMathOperator*{\argmin}{\arg\!\min}
\newcommand{\bm}[1]{\mathbf{#1}}

\begin{document}

\title{Active Image Synthesis for Efficient Labeling}

\author{Jialei~Chen, Yujia~Xie, Kan~Wang, Chuck~Zhang, Mani~A.~Vannan, Ben~Wang, Zhen~Qian%

\IEEEcompsocitemizethanks{
\IEEEcompsocthanksitem J. Chen and C. Zhang are with the H. Milton Stewart School of Industrial and Systems Engineering and the Georgia Tech Manufacturing Institute, Georgia Institute of Technology, Atlanta, GA 30332
USA\protect\\
E-mail: \{jialei.chen, chuck.zhang\}@gatech.edu
\IEEEcompsocthanksitem K. Wang is with the Gerogia Tech Manufacturing Institute, Georgia Institute of Technology, Atlanta, GA 30332
USA\protect\\
E-mail: kwang34@mail.gatech.edu
\IEEEcompsocthanksitem Y. Xie is with the School of Computational Science and Engineering, Georgia Institute of Technology, Atlanta, GA 30332
USA\protect\\
E-mail: xieyujia@gatech.edu
\IEEEcompsocthanksitem B. Wang is with the H. Milton Stewart School of Industrial and Systems Engineering and the School of Materials Science and Engineering, Georgia Institute of Technology and the Gerogia Tech Manufacturing Institute, Atlanta, GA 30332
USA\protect\\
E-mail: ben.wang@gatech.edu
\IEEEcompsocthanksitem M. A. Vannan is with the Marcus Heart Valve Center, Piedmont Heart Institute, Atlanta, GA 30309
USA\protect\\
E-mail: Mani.Vannan@piedmont.org
\IEEEcompsocthanksitem Z. Qian is with the Tencent America, Medical AI Lab, Palo Alto, CA 94306
USA\protect\\
E-mail: qianzhen@tencent.com
}
\thanks{Corresponding author: Jialei Chen and Zhen Qian.}
}

\markboth{IEEE Transactions on Pattern Analysis and Machine Intelligence}%
{Chen \MakeLowercase{\textit{et al.}}: Bare Advanced Demo of IEEEtran.cls for IEEE Computer Society Journals}

\IEEEtitleabstractindextext{%
\begin{abstract}
The great success achieved by deep neural networks attracts increasing attention from the manufacturing and healthcare communities.
However, the limited availability of data and high costs of data collection are the major challenges for the applications in those fields.
We propose in this work AISEL, an active image synthesis method for efficient labeling, to improve the performance of the small-data learning tasks.
Specifically, a complementary AISEL dataset is generated, with labels actively acquired via a physics-based method to incorporate underlining physical knowledge at hand.
An important component of our AISEL method is the bidirectional generative invertible network (GIN), which can extract interpretable features from the training images and generate physically meaningful virtual images. 
Our AISEL method then efficiently samples virtual images not only further exploits the uncertain regions but also explores the entire image space. 
We then discuss the interpretability of GIN both theoretically and experimentally, demonstrating clear visual improvements over the benchmarks.
Finally, we demonstrate the effectiveness of our AISEL framework on aortic stenosis application, in which our method lowers the labeling cost by $90\%$ while achieving a $15\%$ improvement in prediction accuracy.
\end{abstract}

}

\maketitle

\IEEEdisplaynontitleabstractindextext

\IEEEpeerreviewmaketitle

\ifCLASSOPTIONcompsoc
\IEEEraisesectionheading{\section{Introduction}\label{Sec:intro} }
\else
\section{Introduction}\label{Sec:intro} 

\fi

\IEEEPARstart{D}{eep} neural networks (NNs) \cite{mcculloch1943logical,lecun2015deep,bengio2013representation} have achieved superior performance in computer vision tasks \cite{lecun2010convolutional,ImageNet_cvpr09}, and attracts increasing attention from other communities, including manufacturing \cite{wang2018deep} and healthcare \cite{litjens2017survey}.
When fed with a \textit{large} amount of training data (at least in the thousands \cite{du2018many}), NNs have shown great success in extracting high-level features and modeling complex functions.
However, the available data in actual life is often \textit{limited} and \textit{expensive} to collect.
For example, in computer-aided diagnosis of aortic stenosis, a common yet severe heart disease \cite{stewart1997clinical}, doctors are interested in using pre-surgical CT scans to efficiently identify the diseased patients. 
Here, a hospital may only have around a hundred historical records over the years, leading to unsatisfactorily performance for NNs.


In the meantime, thanks to the advances in domain research, underlining physical knowledge is often available for the learning problems in manufacturing and healthcare.
Take the same aortic stenosis application as an example, the pathophysiological reason for the stenosis is mainly due to the deposited calcifications on the valve leaflets and the valve wall, and therefore change the blood flow pattern. The blood flow can be numerically simulated via computational fluid dynamics (CFD, see\cite{anderson1995computational}), using the CT scans as the input geometry and boundary conditions.
Incorporating such knowledge (i.e., simulation) would intuitively improve the learning model since it provides complementary information against the collected historical records.
We present in this paper an \textit{active} sampling method to incorporate underlining physical knowledge via a  \textit{complementary} dataset.

However, there are two major challenges involved in collecting such a complementary dataset. 
First, the inputs of the dataset (i.e., unlabeled images) are difficult to acquire in practice.
This is particularly typical in the medical field, e.g., pre-surgical CT scans, due to clinical, logistic, and economic restrictions. 
Therefore, an effective \textit{synthesis} model for image inputs is needed. 
Second, physical labeling methods are usually expensive.
For example, it may take several hours of computation for a CFD model with complex geometry \cite{anderson1995computational}, and it would be even longer if considering the interaction of blood flow and soft biological tissue \cite{thompson1997general}. 
Within a practical turnaround, one can only afford a relatively \textit{small} amount of labeled experiments. Therefore, an efficient \textit{sampling} strategy is needed for data synthesis.


We propose in this work AISEL (an Active Image Synthesis framework for Efficient Labeling) to actively incorporate the underline physical knowledge in small-data learning.
Our AISEL framework contains two major components.
We first propose the generative invertible network (GIN) -- a novel \textit{bidirectional} image generative model -- to encode the actual images (i.e., the training images) into the defined lower-dimensional feature space, in which candidate virtual images can then be generated. GIN can be viewed as an extension of the generative adversarial networks \cite{goodfellow2014generative} by adding an inverse mapping for feature extraction to the generative mapping.
Moreover, we propose a new uncertainty sampling method to actively select the candidate virtual images in the GIN feature space.
In our sampling method, virtual images are efficiently selected to represent the \textit{distribution} of uncertainty in the energy-distance sense, and therefore both \textit{exploit} the highly uncertain regions and \textit{explore} the entire space without overlap.
Labels for selected virtual images are then obtained via the physical labeling approaches at hand.
By merging the training data and our AISEL dataset, improved downstream models are observed on both toy computer vision/manufacturing applications and the medical application of aortic stenosis.
This paper makes the following contributions:
\begin{enumerate}
\item  We incorporate physical knowledge into the learning process, via a complementary dataset.
This ensures the incorporation of the additional information (by the \textit{physics-based} labeling approaches), and therefore improves the downstream prediction performance.
\item  We propose an efficient image sampling method for complementary dataset. Specifically, it minimizes the predictive uncertainty and mitigates the possible high labeling cost.
\item We propose a new \textit{bidirectional} generative model -- GIN for feature mapping and actively generating virtual images, conditional on the actual images. Noticeable visual improvements compared to the benchmarks are observed. 
\end{enumerate}

The paper is structured as follows. Section 2 summarizes the related works. Section 3 presents the proposed GIN with an emphasis on the difference with GAN. Section 4 discusses the new sampling method and features the whole AISEL learning framework. Section 5 demonstrates the effectiveness of our method in both toy examples and the motivating application of aortic stenosis. Section 6 concludes the work with directions for future research.



\section{Related work}

\textbf{Data augmentation} is widely used for different learning tasks with image inputs \cite{chatfield2014return,perez2017effectiveness}, via image translation, rotation and flip, and changing of the tune and/or brightness to increase the training data size. 
Usually, it assumes such augmentation does not change the label.
However, this may not hold in, e.g., medical images. Taking CT scans as an example, different substances of human tissues correspond to different ranges of image intensity, alterations of which may lead to a completely different interpretation of the pathophysiological condition \cite{chen2018generative}.  
This significantly limits the augmentation methods suitable for manufacturing and healthcare applications.
As to be shown later, the predictive performance with simple augmentation is not good enough.

\textbf{Generative adversarial networks} (GAN) \cite{goodfellow2014generative} opens an era of adversarial training for multiple learning challenges, e.g., image segmentation \cite{zhu2017unpaired} and domain adaptation \cite{isola2017image}. 
We adapt in this work a GAN-based method for the generative model, because (i) compared to variational autoencoder \cite{kingma2013auto,zhao2017towards}, GAN achieves visually better performance, and (ii) compared to generative flow \cite{kingma2018glow}, GAN contains a generative mapping from the low-dimensional features space, which can be used for our new sampling method. 

To achieve efficient image sampling, study design in the feature space is desirable and crucial. However, Most GAN-based methods feature only generation mapping. Exceptions are
adversarially learned inference (ALI) \cite{dumoulin2016adversarially} and bidirectional GAN (BiGAN) \cite{donahue2016adversarial}, which learns both generating mapping and its inverse by a \textit{coupled} architecture of three NNs. The model is proposed mainly for inference and representation learning.
However, complicated architectures and the coupled training of three NNs requires a large amount of data, which is not suitable for our small-data learning problems. Our GIN will be compared with BiGAN to show a noticeable improvement in visual quality.

Conditional GAN (CGAN) \cite{mirza2014conditional} and auxiliary classifier GAN (ACGAN) \cite{odena2017conditional} can generate images with given labels.
Such models can be used to generate \textit{both} virtual images and the corresponding labels for data augmentation \cite{odena2017conditional,frid2018synthetic}.
In our AISEL framework, we \textit{only} generate the input images, while the labels are acquired via physical experiments to incorporate complementary knowledge. 
We will show that the proposed method has noticeable better predictive accuracy compared to the ACGAN-based method.

\textbf{Transfer learning} is another popular approach for small-data learning tasks \cite{yosinski2014transferable,shao2014transfer}. 
Adapting the models trained on natural images (mostly, ImageNet \cite{ImageNet_cvpr09}), researchers are able to fine-tune the pre-trained model coefficients to address the limitation imposed by the small sample size \cite{litjens2017survey}. 
This approach explores the visual cues extracted from natural images and assumes they are also useful in interpreting the training data at hand. 
However, for learning tasks in manufacturing and healthcare, the rationality of such an assumption is unclear. 
For example, comparing CT scans to natural images,  (i) noticeable differences in image appearances are observed, and (ii) pixel intensity value has intrinsically different meanings.
Nevertheless, transfer learning will be served as a baseline for the proposed framework.

\textbf{Active learning} (or sequential experimental design \cite{winer1962statistical} in statistics literature) methods are also used for small-data learning with an oracle labeling method available \cite{settles.tr09,settles2012active}.
They aim to select the next ``good'' input data for labeling. 
Active learning methods are popular in traditional machine learning, with recent improvements for deep learning models \cite{gal2017deep,ducoffe2018adversarial}.
Most active learning methods in the literature assume that a sizeable \textit{unlabeled} dataset is available.
However, in manufacturing and healthcare applications, the unlabeled images are also difficult to acquire in nature.

One of the few exceptions is generative adversarial active learning (GAAL) \cite{zhu2017generative} in literature, which uses GAN to generate unlabeled data. However, GAAL is proposed specifically for the support vector machine classifier. 
Since the support vector machine performs poorly in complicated classification tasks (e.g., our aortic stenosis application), we will compare our method with a modified version of GAAL using a convolutional NN as the classifier.

\textbf{Few-shot learning} is another popular method for small-data learning tasks \cite{wang2019few}. Though it can successfully handle learning tasks with very small training data, it usually requires many such tasks. Here, we only have one task, and therefore we will leave out few-shot learning baselines.

\section{Generative invertible network}
\label{sec:GIN}

In this section, we propose the novel bidirectional GIN as the feature encoding and image generating model, for later efficient image sampling. We first present the GIN architecture with a detailed comparison to GAN. We then show the implementation detail and algorithm for the proposed GIN.

\subsection{Image generating}

\begin{figure}
\centering
\includegraphics[width=0.3\textwidth]{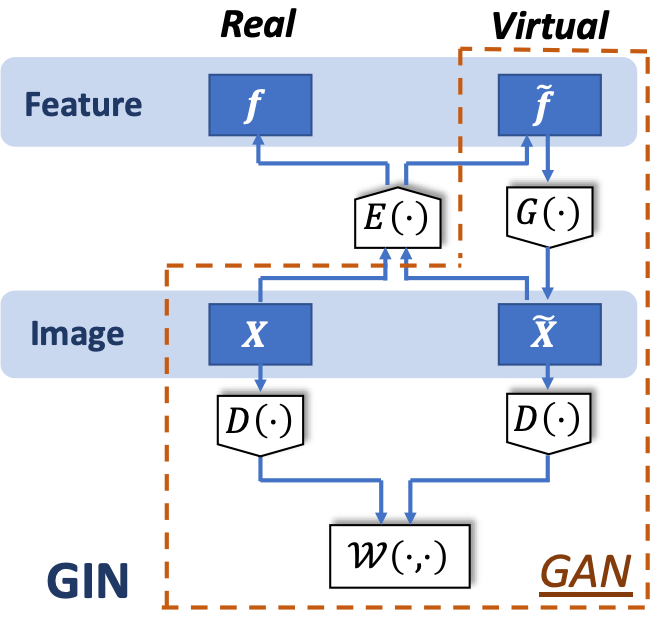}
\caption{\label{fig:GIN} Illustration of the proposed GIN: generator $G(\cdot)$ and discriminator $D(\cdot)$ are obtained by optimizing the Wasserstein distance $\mathcal{W}(\cdot, \cdot)$; encoder $E(\cdot)$ is a sample-to-sample inverse of $G(\cdot)$, explicitly trained by minimizing MSE. Compared to GAN, GIN contains the additional encoder $E(\cdot)$.}
\end{figure}

\label{sec:GAN}

Following the standard GAN \cite{goodfellow2014generative,arjovsky2017wasserstein}, we model the training set images as realizations of the distribution of the images of interest $\mathcal{X}: \mathcal{B}[\mathbb{X}] \mapsto [0,1]$. Here, $\mathbb{X} = \mathbb{R}^{n_1\times n_2}$ denotes the space of images with pixel size $n_1\times n_2$, and $\mathcal{B}[\mathbb{X}]$ is its Borel set \cite{resnick2013probability}. 
Furthermore, in order to efficiently learn the generative mapping and later interpretation, we define a feature space $\mathbb{F} = [-1,1]^r$. Here, $r$ is the pre-defined dimension of the feature space, usually assumed to be much lower than that of the image space.
We set a non-informative, uniform measure $\mathcal{U}$ on the feature space $\mathbb{F}$, which represents the lack of understanding of the feature space. 
The goal is to learn a generative mapping $G(\cdot): \mathbb{F} \mapsto \mathbb{X}$ which best pushforwards the uniform measure $\mathcal{X}'=G_\#(\mathcal{U})$ and mimics the target measure $\mathcal{X}$. 
We use in this work the Wasserstein-1 metric \cite{villani2008optimal} as the loss function:
\begin{equation}
\mathcal{W}(\mathcal{X}, \mathcal{X}') = \inf_{\gamma} \int _{\mathbb{X}\times \mathbb{X}}  \|\bm{X}-{\bm{X}'}\|_2 d \gamma(\bm{X},{\bm{X}'}),
\end{equation}
where $\|\cdot\|_2$ is the $l_2$ norm, and the infimum is obtained with respect to all the possible joint distribution $\gamma: \mathcal{B}[\mathbb{X}\times\mathbb{X} ]\mapsto [0,1]$ whose marginals are $\mathcal{X}$ and $\mathcal{X}'$.
We adopt the Kantorovich-Rubinstein dual form \cite{villani2008optimal} of Wasserstein distance for efficient implementation:
\begin{equation}
\mathcal{W}(\mathcal{X}, \mathcal{X}')= \sup_{\|D(\cdot)\|_L\leq 1} \mathbb{E}_{x\sim \mathcal{X}}[D(x)] - \mathbb{E}_{x'\sim \mathcal{X}'}[D(x')].
\end{equation}
Here $D(\cdot): \mathbb{X}\mapsto \mathbb{R}$ is an evaluating function and $\|D(\cdot)\|_L\leq 1$ represents that $D(\cdot)$ is Lipschitz-1 continuous \cite{arjovsky2017wasserstein}.

We use a NN to approximate the generating mapping $G(\cdot)$, named \textit{generator}, and another NN for the evaluating function $D(\cdot)$, named \textit{discriminator}. The aim is to find the optimum of the following minimax function:
\begin{equation}
\min_{G(\cdot)} \max_{D(\cdot)} \mathbb{E}_{x\sim \mathcal{X}_n}[D(x)] - \mathbb{E}_{u\sim \mathcal{U}}[D(G(u))],
\label{equ:trainGIN}
\end{equation}
where $\mathcal{X}_n$ is the empirical measure for the training images with size $n$, and $ \mathcal{U}$ is the uniform measure to be pushforwarded. Iterative training strategy can be adapted. Further discussion on the numerical implementation and the convergence analysis can be found in Section \ref{Sec:Alg1}.

\subsection{Feature encoding} 
\label{Sec:Encoder}
Assume for now the generating mapping $G(\cdot)$ is known. We are interested in finding an encoding mapping $E(\cdot): \mathbb{X} \mapsto \mathbb{F}$ to embed the images back to the feature space, which, to be shown in Theorem \ref{Thm:Inverse}, is an inverse of $G(\cdot)$. 
Similar to the generating mapping, we use a NN to parametrize $E(\cdot)$, named \textit{encoder}. 
Since the task here is to extract the feature vectors from images, a convolutional neural network (CNN) is used with mean square error (MSE) loss: 
\begin{equation}
   \min_{E(\cdot)} \mathbb{E}_{\bm{X} \sim \mathcal{X}_n}\|E(\bm{X})-f\|_2^2,
    \label{equ:actualMSE}
\end{equation}
where $f\in \mathbb{F}$ is the corresponding feature vector associated with the image $\bm{X}$. The reason we use an MSE loss is due to the desired regression task here: we want a strong metric to ensure the sample-to-sample inverse of $G(\cdot)$. Furthermore, we want $E(\cdot)$ dedicated only on this inversion task, and therefore permits an efficient sampling method later in Section 4.1.

However, the difficulty is that the feature $f$ for the actual image $\bm{X}$ is unknown.
In other words, $E(\cdot)$ cannot be learned from the dataset of actual images at hand. Therefore, we revise \eqref{equ:actualMSE} to
\begin{equation}
\label{Equ:TrainCNN}
    E(\cdot) = \argmin_{E(\cdot)} \mathbb{E}_{u \sim \mathcal{U}}\|E(G(u))-u\|_2^2,
\end{equation}
where $\bm{X}' = G(u)$ is the generated virtual images. 
Another advantage of using the virtual data points 
is that the data size of the virtual images can be large, since one may generate as many virtual images $\bm{X}'$ as needed. 
We expect the encoder $E(\cdot)$ learned via \eqref{Equ:TrainCNN} using \textit{virtual} data points (instead of the \textit{actual} images in training set) is still the inverse of the generator $G(\cdot)$. Formally, we have the following Theorem:

\begin{theorem}
Denote the target distribution measuring as $\mathcal{X}$ on image space $\{\mathbb{X},\mathcal{B}[\mathbb{X}]\}$. Assume the generator $G(\cdot)$ is obtained by \eqref{equ:trainGIN} with the training error $<\epsilon$ and encoder $E(\cdot)$ is obtained by \eqref{Equ:TrainCNN} with the training error $< \delta$. If both $G(\cdot)$ and $E(\cdot)$ are Lipschitz-$L$ continues, then the reconstruction error $\mathbb{E}_{x \sim \mathcal{X}}[G(E(x))-x]^2$ can be bounded by $(L^2+L+1)\epsilon+L\delta$.
\label{Thm:Inverse}
\end{theorem}

\noindent This means the obtained $G(\cdot)$ and $E(\cdot)$ are inverses of each other in the sense of minimizing the reconstruction error. The proof of this theorem can be found in Appendix \ref{App:thm1}.

The reason for introducing the encoding mapping $E(\cdot)$ as the inverse of generating mapping $G(\cdot)$ is twofold. 
First, we can use $E(\cdot)$ to encode the actual images as vectors in the feature space $\mathbb{F}$.
They can then be used as lighthouses in $\mathbb{F}$, and provide intuitive understanding of the feature space (we will discuss this later in Section 5.2.2).
Second and perhaps more important, in the following sampling method, we want to sample virtual images for better predictive performance with a limited labeling budget. In our AISEL method (see Section \ref{sec:Doe}), the sampling is performed in $\mathbb{F}$ rather than the image space $\mathbb{X}$, for its lower dimension and the physical meaning. Moreover, while sampling virtual images, we need guidance from the features of the actual images. For example, one may not want to sample images that are too similar to any of the actual images to better explore the whole $\mathbb{F}$.
This can be achieved by introducing a separating distance between virtual images and actual images (we will come back to this in Section 4.1); this needs to encode the actual images to the feature space by $E(\cdot)$.

\subsection{Summary and algorithm for GIN} 
\label{Sec:Alg1}
Putting everything together, the proposed GIN consists of three NNs: a generator $G(\cdot)$ for generating virtual images, an encoder $E(\cdot)$ for feature embedding, and a discriminator $D(\cdot)$ for computing the Wasserstein distance. Fig.\ref{fig:GIN} illustrates the architecture of GIN.
Note that in GIN, $G(\cdot)$ and $E(\cdot)$ is decoupled due to the limited training data. 
We present Algorithm \ref{Alg:GIN} to train the proposed GIN.
The first part of the algorithm is to train a generator $G(\cdot)$ parameterized by $\theta$, and the second part is to train an encoder $E(\cdot)$ parameterized by $\gamma$. 
The generator and discriminator are coupled trained as GAN, while the additional encoder is separately trained by the virtual images sampled by $G(\cdot)$.
In the small-data situation, the proposed GIN along with the associated algorithm can achieve visual improvement in practice, compared to other methods like BiGAN; we will provide a detailed discussion in Section 5.1.2.

\begin{algorithm} [t!]
\caption{Generative invertible network\label{Alg:GIN}}
\begin{algorithmic}[1]
\Procedure{GIN}{$\{\bm{X}_i\}_{i=1}^n$}
\State Initialize $G_\theta(\cdot)$, $D_w(\cdot)$, and $E_\gamma(\cdot)$
\While {$\theta$ has not converged}
\State Sample $\{f'_i\}_{i=1}^m \sim \mathcal{U}$
\State $L_G = -\sum_{i=1}^m D_w(G_\theta (f'_i))$
\State $\theta = \theta - \alpha \nabla L_G$ \Comment{Train generator}
\State $G(\cdot)= G_\theta(\cdot)$
\For {$t=0,\cdots,n_d$}
\State Sample $\{\bm{X}_i\}_{i=1}^m$ a batch from the actual data.
\State Sample $\{f'_i\}_{i=1}^m \sim \mathcal{U}$
\State $L_D = \sum_{i=1}^m D_w(\bm{X}_i) - \sum_{i=1}^m D_w(G_{\theta}(f'_i))$
\State $w=w+\alpha \nabla L_D$ \Comment{Train discriminator}
\State $w = \text{clip} (w, -\beta, \beta)$
\EndFor
\EndWhile
\While {$\gamma$ has not converged}
\State Sample $\{f'_i\}_{i=1}^m \sim \mathcal{U}$
\State Generate $\{\bm{X}'_i=G(f'_i)\}_{i=1}^m \sim \mathcal{X}'$
\State $L_E =\sum_{i=1}^m (E_\beta(\bm{X}'_i)-f'_i)^2$
\State $\gamma = \gamma - \alpha \nabla L_E$ \Comment{Train encoder}
\State $E(\cdot)= E_\gamma(\cdot)$
\EndWhile
\State \textbf{return} $G(\cdot)$, $E(\cdot)$
\EndProcedure
\end{algorithmic}
\end{algorithm}

One may be interested in finding out how ``real'' the virtual images can be generated using the proposed Algorithm \ref{Alg:GIN}, since multiple heuristic strategies are involved (e.g., iterative training of $D(\cdot)$ and $G(\cdot)$, and clip). Furthermore, note that the above computation is done with \textit{samples} of actual images, i.e., the \textit{empirical} probability measure $\mathcal{X}_n$, instead of the original probability measure $\mathcal{X}$. Therefore, we have the following theorem for asymptotic convergence.
\begin{theorem}
Denote the target measure as $\mathcal{X}$ and its empirical measure represented by the training set data as $\mathcal{X}_n$. Assuming both neural networks $G(\cdot)$ and $D(\cdot)$ are obtained as the optimum of target function (\ref{equ:trainGIN}).
Let $\mathcal{X}'$ be the measure obtained by the proposed Algorithm \ref{Alg:GIN}. Specifically, it is a pushforwarded measure of $\mathcal{U}$ by $G(\cdot)$, i.e., $\mathcal{X}'[S]=(G_\#(\mathcal{U}))[S]=\mathcal{U}[G^{-1}(S)]$ for any $ S \in \mathcal{B}[\mathbb{X}]$. As the training data size approaches infinity, we have $\mathcal{X}' \to \mathcal{X}$ in distribution.
\label{Thm:SameDis}
\end{theorem}
\noindent
The proof, following \cite{arjovsky2017wasserstein}, can be found in Appendix \ref{App:thm2}.

Theorem \ref{Thm:SameDis} suggests that, if we have enough training data, the generated images are \textit{real} enough compared to the actual images. 
Specifically, it means the generated images and the measure $\mathcal{X}'$ have the following two properties. 
First and most importantly, the supports of the two measures are the same, i.e., $supp(\mathcal{X}') = supp(\mathcal{X})$, with probability 1.0. This is a natural corollary of Theorem \ref{Thm:SameDis}. It means any generated virtual image $\bm{X}'$ can be regarded as a draw from the measure of actual images $\mathcal{X}$, i.e., $p_{\mathcal{X}}(\bm{X}')>0$, where $p_{\mathcal{X}}(\cdot)$ denotes the probability density of $\mathcal{X}$. In other words, the generated images are always physically meaningful. 
Moreover, besides their support, the two probability measures themselves are the same asymptotically. This means the probability of generating the same group of images (e.g., CT scans of male patients, or CT scans of patients with no complications) is the same, which is an implicit requirement when endowing the feature space with physical meaning and for the following sampling method. Though in practice we are dealing with a small-data situation, it is still appealing to have this asymptotic convergence property.

\section{AISEL Framework}

We present now the proposed AISEL framework for small-data problems. 
For the simplicity of illustration, we assume the learning task is a \textit{classification} problem with images inputs (this is the case of the motivating application);
the proposed framework can be easily extended to regression tasks, which will not be elaborated on in this paper.

We adopt here the standard $K$-class classification setting, which uses input images $\bm{X}_i \in \mathbb{X}$ to predict the probability of assigning to each class $y_i \in [0,1]^K$.
The native classifier $C(\cdot):\mathbb{X} \mapsto [0,1]^K$, parameterized by a NN, refers to the model learned with \textit{only} the small training data at hand. 
With the native model $C(\cdot)$ and GIN (i.e., generator $G(\cdot)$ and encoder $E(\cdot)$) at hand, we first propose the new sampling method to select $m$ virtual images. We then discuss the physical labeling methods and why they are crucial in improving performance. Finally, an algorithmic framework is presented for implementation.

\subsection{Active image sampling}
\label{sec:Doe}

We start with using the entropy \cite{shannon1948mathematical} to quantify the uncertainty of the native model $C(\cdot)$. For any input image $\bm{X}_0 \in \mathbb{X}$ and the corresponding predicted label $y_0=C(\bm{X}_0)$, we have
\begin{equation}
\label{equ:entropy1}
H(\bm{X}_0) = -\sum_{k=1}^K y_0[k] \log(y_0[k]),
\end{equation}
where, $y_0 = [y_0[1],y_0[2],\cdots, y_0[K]]^T$. 
The reason for using entropy to quantify uncertainty can be explained as: (i) If we are sure about the class label of the input image, e.g., $y_0[1]=1$ and $y_0[k]=0, k=2,\cdots,K$; the corresponding entropy is zero, meaning no uncertainty exists. (ii) Consider another extreme situation that $y_0[k]=1/K, k=1,\cdots, K$. One can easily check this maximizes the entropy, reflecting the maximal uncertainty for the label of that image.

The image space $\mathbb{X}=\mathbb{R}^{n_1\times n_2}$ is too high dimensional to handle in reality. Since GIN is already obtained, we can measure the uncertainty (i.e., entropy), for any $f_0\in \mathbb{F}$ in the \textit{feature} space:
\begin{equation}
\label{equ:entropy}
h(f_0) = H(G(f_0)) = -\sum_{k=1}^K E(G(f_0))[k] \log(E(G(f_0))[k]).
\end{equation}
Here, we select the \textit{features} of the complementary dataset in the feature space $\mathbb{F}$, rather than in the high-dimensional image space $\mathbb{X}$.
Besides the dimensionality, our $G(\cdot)$ can capture the intrinsic structure of the image space $\mathbb{X}$ -- selecting features in $\mathbb{F}$ (and then generating images via $G(\cdot)$) can ensure the existence of physical meaning. This is because any generated images $G(f_0)$ with any $f_0 \in \mathbb{F}$ is physically meaningful thanks to the $G(\cdot)$, while randomly sampled $\bm{X}_0 \in \mathbb{X}$ is most likely a matrix without any visual clue.

The entropy $h(\cdot): \mathbb{F}\mapsto [0,\log K]$ can also be viewed as a (unnormalized) probability density on the measurable space $\{\mathbb{F},\mathcal{B}[ \mathbb{F}]\}$. We denote this \textit{uncertainty measure} as $\mu_h$. 

We then propose to select the best set of $m$ virtual images, by matching its empirical distribution to the uncertainty measure:
\begin{equation}
\label{equ:minDist}
f_{1:m}' = \argmin_{f_{1:m}'} dist(\mathcal{F}'_m,\mu_h).
\end{equation}
Here, $dist(\cdot,\cdot)$ is a distance metric, $f_{1:m}'=\{f_i'\}_{i=1}^m$ denote the selected features, and $\mathcal{F}'_m$ denotes the empirical measure for $f_{1:m}'$.
Intuitively, \eqref{equ:minDist} means to assign more points to higher uncertainty regions (of the native model), and therefore exploit those regions. 
Furthermore, if taking the Bayesian perspective, it can be viewed as changing the initial uniform distribution, i.e., the non-informative prior, to the posterior distribution of uncertainty given the actual training dataset.

Motivated by the literature in the statistical community \cite{joseph2015sequential,mak2018support}, we select the energy distance as the metric $dist(\cdot,\cdot)$ between distributions. Therefore, we minimize:
\begin{equation}
\label{equ:sp}
\min_{f_{1:m}'}\sum_{i=1}^m \mathbb{E}_{\gamma \sim \mu_h}\|f'_i-\gamma\|_2 -\frac{1}{2m}\sum_{i=1}^m\sum_{j=1}^m \|f'_i-f'_j\|_2.
\end{equation}
Note again, the above sampling optimization is conducted in the feature space. We observe from \eqref{equ:sp} that the selected features not only try to match the target uncertainty measure in the expectation sense (the first term), but also separate from one another (the second term). The separating property is of great importance; this is because any two selected features that are too close to each other can be viewed as a waste of the expensive labeling process.

Furthermore, the selected features should also be separated from the features of the \textit{actual} images, again to avoid waste. This can be taken into account by the following modification of \eqref{equ:sp}:
\begin{equation}
\label{equ:seq_sp}
\min_{f_{1:m}'} \sum_{i=1}^{m} \mathbb{E}_{\gamma \sim \mu_h}\|f'_i-\gamma \|_2 -\sum_{i=1}^{m+n}\sum_{j=1}^{m+n}\frac{\|f'_i-f'_j\|_2}{2(m+n)},
\end{equation}
where $n$ is the size of the actual dataset and let $f'_i=f_i$ for actual images with indies $i= m+1, \cdots, m+n$. Comparing \eqref{equ:seq_sp} to \eqref{equ:sp}, we notice the difference lies in the second term, where the separating property is incorporated not only between the selected features but also between the selected features and the features of actual images. 
We use \eqref{equ:seq_sp} for sampling features, and then generate an AISEL dataset via $G(\cdot)$. The following theorem ensures the generated AISEL dataset follows the target uncertainty measure in distribution.

\begin{theorem}
Let target uncertainty measure be $\mu_{H}$ in \eqref{equ:entropy1} and the selected features by \eqref{equ:seq_sp} be $f'_{1:m}$ with size $m$. Assume the $G(\cdot)$ is continuous. Further denote the set of virtual images $\{\bm{X}'_i\}_{i=1}^m=\{G(f'_i)\}_{i=1}^m$, with its empirical measure $\mathcal{X}'_m$. We then have  $\mathcal{X}'_m \to \mu_{H}$ in distribution.
\label{Prop:SP}
\end{theorem}
\noindent Here, we show the convergence in the distribution of the \textit{images} (rather than the \textit{features}), as the images are the quantity of interest. Therefore, a continuous assumption on $G(\cdot)$ is needed according to continuous mapping theorem.
The proof can be found in Appendix \ref{App:thm4}; it follows from \cite{mak2018support}. 

The proposed sampling strategy \eqref{equ:seq_sp} reveals an important trade-off. Consider the first term, where the selected features are forced to be close to the target uncertainty measure. Since the density of our target measure \eqref{equ:entropy1} is high when the uncertainty is high, the selected features can be viewed as exploiting the highly uncertain regions. 
Now consider the second term, where the separating distance is maximized. This suggests the selected features should be away from (i) one another and (ii) the features for actual images. Therefore, selected features are forced to be spread out and fill the whole feature space -- they explore the entire feature space. Putting both parts together, selected features for virtual images jointly exploit the highly uncertain regions and explore the entire image space. This trade-off of our AISEL dataset will be shown as the key to improve the classification performance.

We want to make a few remarks here.
First, the proposed sampling method, specifically the uncertainty measure \eqref{equ:entropy}, is specifically for the classification problem at hand. 
With a different uncertainty measure, the proposed approach is also suitable for regression problems. For example, one may obtain the measure via predictive variance using kernel regression \cite{nadaraya1964estimating} or kriging \cite{matheron1963principles} methods. 
Second, our method is motivated by active learning literature \cite{settles.tr09}, where the next input is selected from a pool of candidates with maximal uncertainty. Different from those methods, our method (i) conducts sampling in a much lower-dimensional GIN feature space due to the intrinsic structure of image space and computational efficiency and (ii) sample a batch of images for labeling to both explore and exploit the design space.
Moreover, it is worth pointing out that the proposed method is possible only when \textit{both} the generating mapping $G(\cdot): \mathbb{F}\mapsto\mathbb{X}$ and the encoding mapping $E(\cdot): \mathbb{X}\mapsto\mathbb{F}$ are available via GIN. In particular, $G(\cdot)$ is used to generating images based on the selected features, while $E(\cdot)$ is used to embed the features of the actual images to guide the sampling.
That is the key reason why we propose a \textit{bidirectional} GIN in Section 3.
Last but not least, the proposed method can also be used to balance the label distribution with a modification of our uncertainty measure \eqref{equ:entropy}. See Appendix \ref{app:balance} for more discussion.

\begin{figure}[t]
\centering
\includegraphics[width=0.49\textwidth]{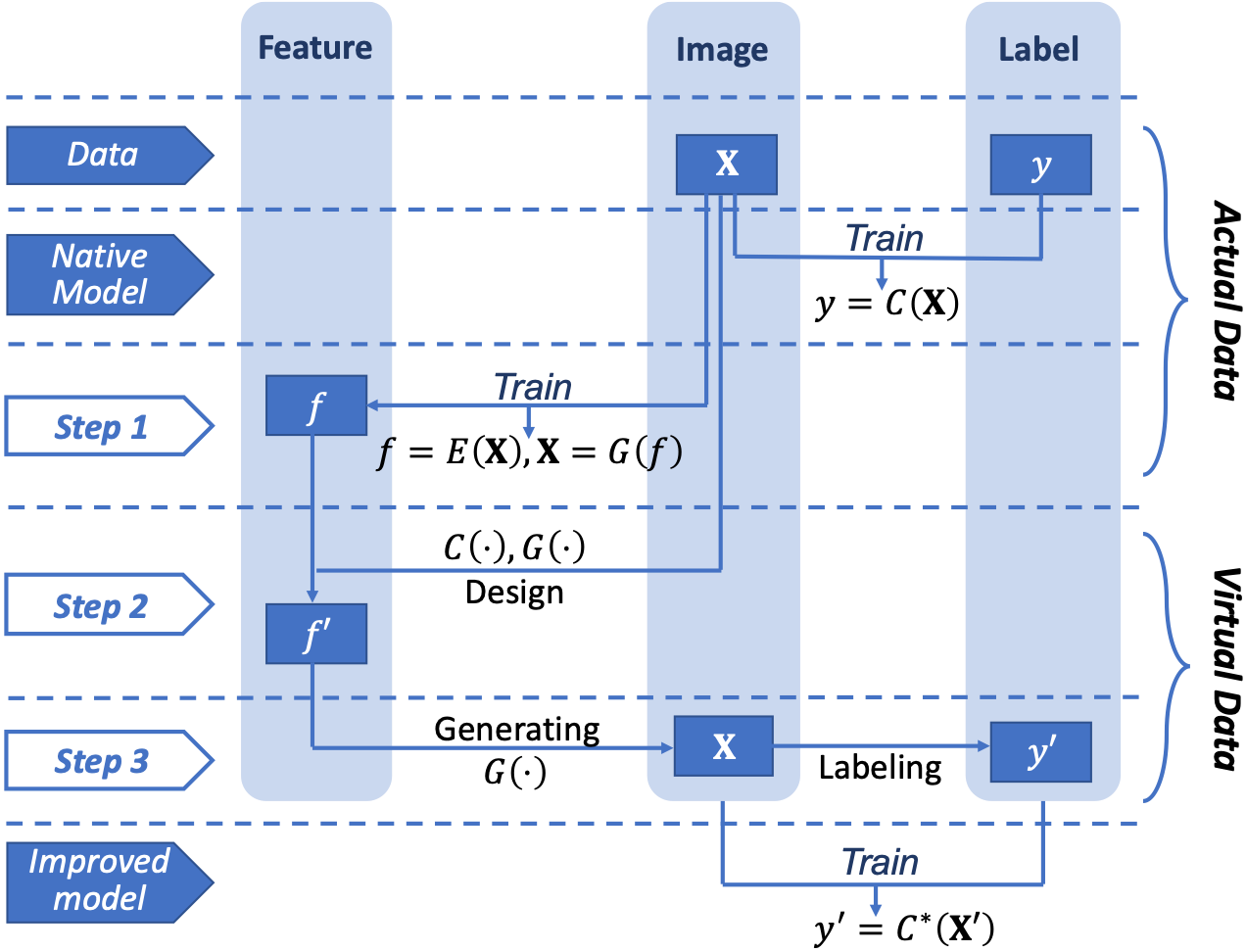}
\caption{\label{fig:Meth} The proposed three-step framework AISEL to efficiently sample AISEL dataset and improve classification.}

\end{figure}

\subsection{Labeling by physical principles}
\label{Sec:PL}
A key component of the proposed AISEL framework, different from data augmentation, is the incorporation of \textit{physical knowledge} while learning.
This is due to the circumstances of real-world applications in manufacturing and healthcare:
(i) the size of the historical records is small, leading to a poor learning model; and 
(ii) thanks to the advances in domain research, physical knowledge is oftentimes available yet expensive in implementation.
Therefore, we want to build a bridge to efficiently combine both the historical data (via the learning model) and physical knowledge (via physical labeling). 
The resulting model can be viewed as one that has learned from data and been taught by physical knowledge, and therefore better performance can be achieved.

Here, we \textit{efficiently} incorporate physical knowledge via a complementary AISEL dataset.
Specifically, we separately acquire the input image and the output label.
For virtual images, \eqref{equ:seq_sp} is used to efficiently sample a set of features to minimize the predictive uncertainty, and GIN is then used to map those features to images.
Meanwhile, for the labels, we use the physical labeling method at hand.
We then combine the actual dataset and AISEL dataset to learn the downstream classification model. 
With the proposed uncertainty sampling method, our AISEL dataset (i) contains complementary information from physical knowledge, and (ii) efficiently exploit and explore the image space, and therefore improves the downstream learning performance.

Lots of different physics-based labeling approaches are available.
For example, finite element analysis \cite{zienkiewicz1977finite} and computational fluid dynamics \cite{anderson1995computational} can solve partial differential equations (i.e., representation of physical knowledge) numerically. These methods can be used to label the input images in, e.g., manufacturing applications.
Physical experiments can also be applied.
With the advances in additive manufacturing, tissue-mimicking 3D printing \cite{chen2018efficient} with an in-vitro study \cite{qian2017quantitative} can be used for medicine-related learning tasks.
If none of the above exists, one can also consult the experts or use certain empirical physical relationships; these methods can be used in computer vision problems.
The specific approach should be made on a case-by-case basis, with the available resources at hand. 

It is important to note that labeling one input image via physical-based approaches is usually \textit{expensive}. 
For example, in medicine-related applications, it may take several hours of computation for a CFD model with complex geometry \cite{anderson1995computational}, and it would be even longer if considering the interaction of blood flow and soft biological tissue \cite{thompson1997general}. 
This is one of the reasons for introducing an efficient and effective sampling method to design our AISEL dataset. 
Viewed this way, our approach can also be used to address the problem where an expensive simulator available, and we want to use that simulator actively for the classification tasks.

\subsection{Summary of the AISEL framework}

\begin{algorithm} [t!]
\caption{Improving classification by AISEL framework}\label{Alg:Overall}
\begin{algorithmic}[1]
\State \quad \quad  \textbf{Native model}  
\State Train CNN, $C(\cdot)$ = CNN($\{\bm{X}_i,y_i\}_{i=1}^n$)
\State \quad \quad  \textbf{Step 1: Train GIN}  
\State Set the feature space $\mathbb{F}=[-1,1]^r$
\State Set prior uniform measure on $\mathbb{F}$
\State Train $G(\cdot)$,$E(\cdot)$ = GIN($\{\bm{X}_i\}_{i=1}^n$) by Algorithm \ref{Alg:GIN}
\State \quad \quad  \textbf{Step 2: Sampling features}  
\State Obtain the uncertainty measure $\mu_h$ by \eqref{equ:entropy}
\State Obtain features for actual images, ${f}_i=E(\bm{X}_i)$
\State Optimize \eqref{equ:seq_sp} by CCP and obtain features ${f}'_j$
\State \quad \quad  \textbf{Step 3: Acquiring AISEL dataset}  
\State Generate actual images, $\bm{X}'_j=G(f'_j)$, $j=1,2,...,m$
\State Obtain labels, $y'_j$ of $\bm{X}'_j$ by physical approaches
\State \quad \quad  \textbf{Improved model}  
\State Train CNN, $C^*(\cdot)$ =CNN($\{[\bm{X}_i,\bm{X}'_j],[y_i,y'_j]\}$)
\end{algorithmic}
\end{algorithm}

In summary, we propose an AISEL framework to efficiently incorporate physical knowledge at hand and improve the classification performance. 
The native model $C(\cdot)$ can be first obtained using the small training data.
As illustrated in Fig.\ref{fig:Meth}, our AISEL framework contains three steps.
First, the proposed GIN is trained using the actual images,  providing a feature space $\mathbb{F}$, and bidirectional mappings between it and the image space (i.e., the generating mapping $G(\cdot)$ and the encoding mapping $E(\cdot)$).
Second, the uncertainty of $C(\cdot)$ at different locations in $\mathbb{F}$ is quantified via entropy, and then the features for virtual images are sampled via \eqref{equ:seq_sp}. 
Third, virtual images are generated by $G(\cdot)$, and then labeled by the physics-based approach.
Finally, the additional AISEL dataset is merged to the original training set, and an improved classifier $C^*(\cdot)$ can be trained.
With the proposed AISEL framework to actively incorporate complementary knowledge via labeling, we will show later in the experiments, $C^*(\cdot)$ can achieve better classification accuracy.

We propose Algorithm \ref{Alg:Overall} for our AISEL framework. In our implementation, the native model and improved model are parameterized by CNN, for its popularity in the image classification tasks. 
Other, perhaps more advanced architecture (e.g., ResNet \cite{he2016deep}) can also be used.
The optimization of \eqref{equ:seq_sp} can efficiently implement by the convex-concave procedure (CCP, see \cite{mak2018support}).
Note that for all the NNs, especially the native model and the GIN, data augmentation methods (e.g., rotation and horizontal flip) are used.
Note that our method can also be used for sequential implementation -- run Algorithm \ref{Alg:Overall} interactively to generate a series of AISEL datasets and therefore provide even better improvement, if budget allows.  

\section{Experiments}
\label{sec:exp}
We first conduct toy computer vision experiments, and provide more insights on our AISEL framework. We then deploy the proposed method to the medical application of aortic stenosis, with emphasis on the pathophysiological meaning of the proposed framework. 

\subsection{Toy computer vision applications}

We conduct experiments on small versions (400 in total for training) of two single-channel (i.e., grayscale) computer vision datasets -- Fashion \cite{xiao2017/online} and MNIST \cite{lecun1998mnist}. 
The two datasets are of particular interest due to their visual similarity to the images in the manufacturing process and modeling.
For example, the images captured by a thermal camera (or simulated via finite element analysis), representing a gray-level temperature contour, can be used to predict the throughput in steel manufacture and conduct quality control in semiconductor manufacturing \cite{chen2019adaptive}.
In the following subsections, we illustrate the visual performance of the proposed GIN, and the improvement in classification by our AISEL method, \textit{only} on the Fashion dataset; similar observation also applies for MNIST (see Appendix \ref{app:MNIST}). 

\begin{figure}[]
\centering
\includegraphics[width=0.49\textwidth]{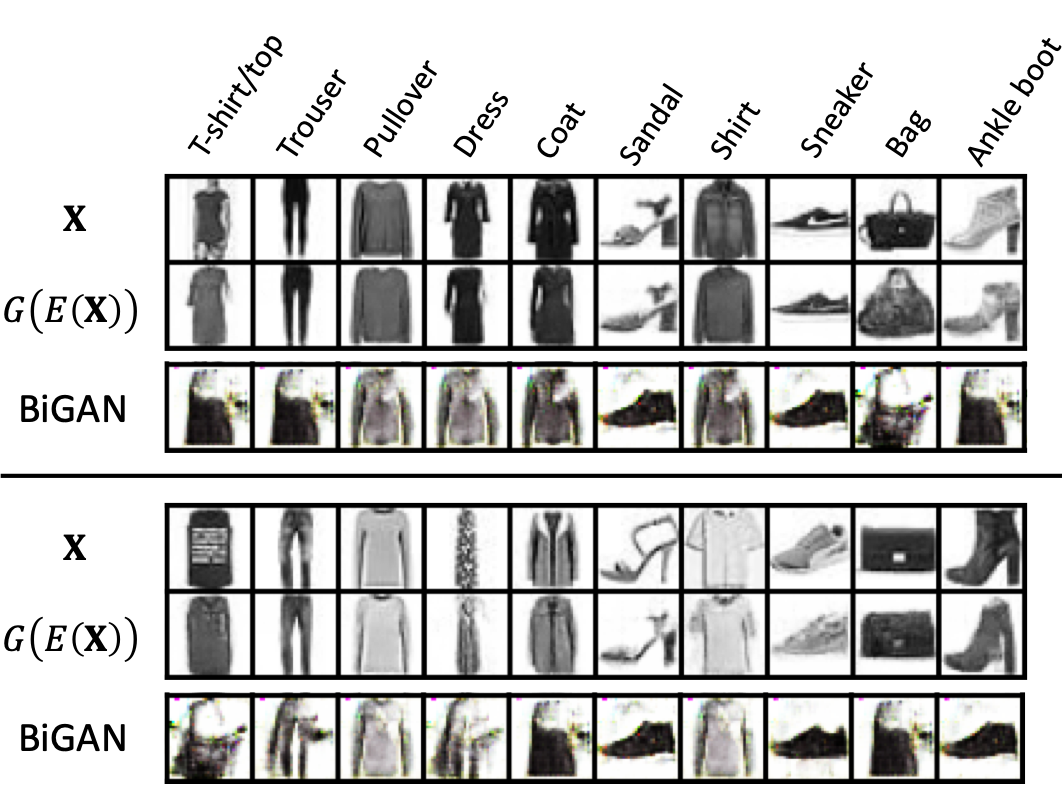}
\caption{\label{fig:SynthRe} Qualitative results for our GIN on Fashion data, including the training data $\bm{X}$ of all ten classes, our reconstructions $G(E(\bm{X}))$ and reconstructions via BiGAN \cite{donahue2016adversarial}.}
\end{figure}

\subsubsection{Fashion dataset}
The Fashion dataset \cite{xiao2017/online} is an MNIST-like dataset of Zalando's article images. As shown in Fig.\ref{fig:SynthRe} (see the rows $\bm{X}$), it contains ten classes of outfits.  We observe that the images associated with classes ``T-shirt/top'', ``coat" and ``shirt" are visually similar in nature, resulting in a more challenging classification task than MNIST. Lots of works have been dedicated to classifying the Fashion dataset \cite{fashionGit}, and the leading accuracy is $96.7\%$ by WideResNets \cite{zagoruyko2016wide}.
We use this model as our labeling approach (see Section 5.1.3 for a detailed discussion).

The original Fashion dataset contains a large amount of training data (60,000 in total). To mimic the real small-data situation in manufacturing, we randomly sample 400 as our training set, with roughly 40 data per label. The original test set (10,000 in total) remains untouched.

\label{Sec:SynData}

\subsubsection{Visual results of GIN}
\label{Sec:SynGIN}
We test first the proposed GIN. 
The dimension of the feature space is set to be $r=2$, i.e., $\mathbb{F}=[-1,1]^2$. This is only for visualization purposes; for actual employment (and in the later application of aortic stenosis), we suggest using a higher $r$ for better performance. 
The detailed architecture of the three NNs can be found in Appendix \ref{App:imp}.
Fig.\ref{fig:SynthRe} shows the generated images (see the rows $G(E(\bm{X}))$). Visually, they look sharp and reasonable without apparent mode dropping.

\textbf{Reconstruction test.}
To visually show the encoder $E(\cdot)$ is the inverse of generator $G(\cdot)$, we conduct the following reconstruction test: for any actual image $\bm{X}_i$, its feature is extracted $f_i=E(\bm{X}_i)$, and then a reconstructed image is generated based on that feature $G(f_i)=G(E(\bm{X}_i))$, which is compared with the actual $\bm{X}_i$ visually.
The test results of all ten classes are shown in Fig.\ref{fig:SynthRe}, with $\bm{X}$ denoting the actual images, and  $G(E(\bm{X}))$ denoting the reconstructing ones. The similarity between the two is noticeable, especially in the sense of the same class. Note that we have already proven that $G(\cdot)$ and $E(\cdot)$ are inverses of each other in an ideal situation (see Theorem \ref{Thm:Inverse}), and here we show the inverse can be achieved in practice.


\begin{figure}[]
\centering
\includegraphics[width=0.49\textwidth]{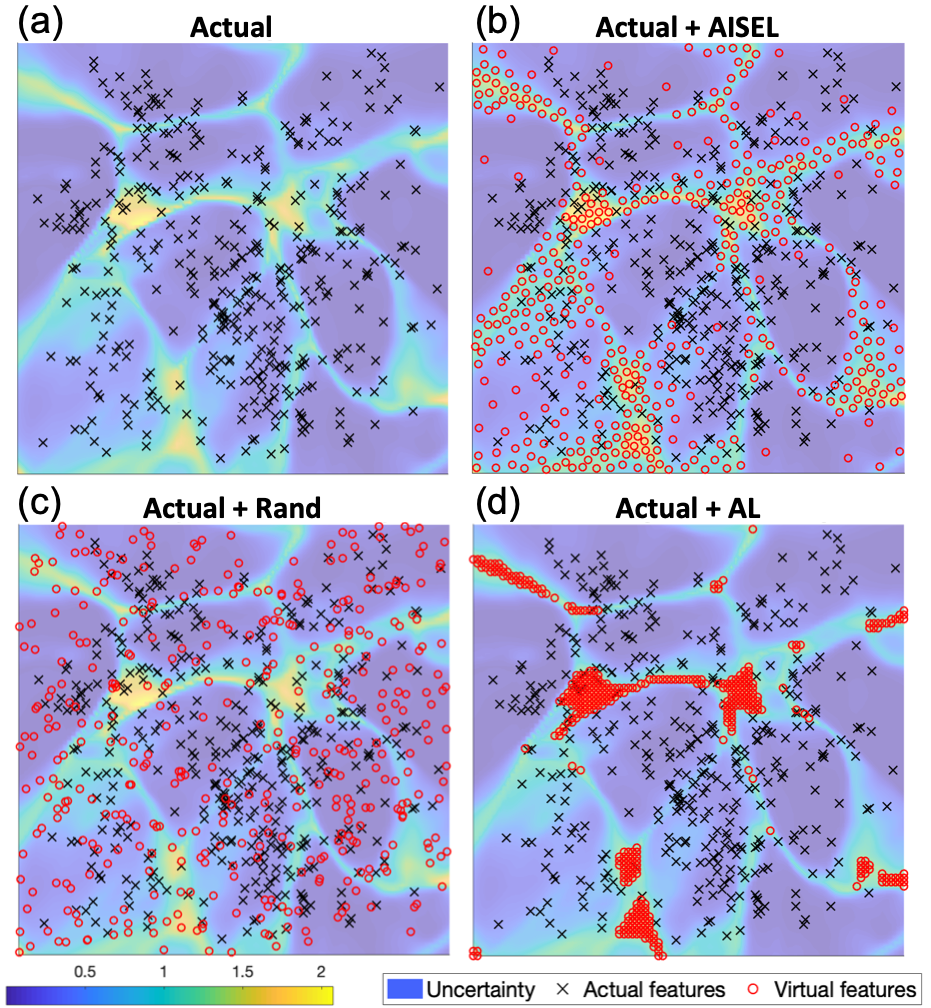}
\caption{\label{fig:SynPredict} 
A comparison of the selected features by our AISEL method, the random sampling method and the active learning method, with uncertainty measure \eqref{equ:entropy} as background.}
\end{figure}

\begin{table*}[]
\centering
\caption{The classification accuracy applying our AISEL method and baselines, on the Fashion dataset and MNIST.
}
\label{tab:Syn}
\begin{tabular}{c|c|cc|cccc|cc}
\toprule
 & \textbf{Native (400)} &{Transfer} &{ACGAN}  & {Rand (+400)}  & {Rand (+5000)} & \textbf{AISEL (+400)}  & {AL (+400)}  &{Oracle (800)} &{Oracle (all)} \\ \hline
\textit{Fashion}          & \textbf{72.8\%} & 74.3\%    & 72.3\%       & 76.4\%     & 80.7\%     & \textbf{81.9\% } & 78.2\%  & 81.3\%  & 96.7\%   \\ 
\textit{MNIST}        & \textbf{88.2\%} & 87.4\%     & 85.9\%     & 90.2\%  & 91.6\%       & \textbf{91.2\%}     & 90.4\% & 91.3\% &  99.2\%   \\ 
\toprule
\end{tabular}
\end{table*}

\textbf{Comparison with BiGAN.} We compare our GIN with BiGAN in literature, which also features a bidirectional mapping \cite{donahue2016adversarial}.
The architecture of BiGAN is set to be similar to GIN, with the same feature dimension $r=2$ and hidden layers.
Fig.\ref{fig:SynthRe} (see the rows ``BiGAN'') shows the reconstruction test conducted by BiGAN using the same set of actual images $\bm{X}$ as GIN. 
Further tuning (e.g., learning rate and hidden dimensions) of the BiGAN is also conducted, with similar performance (also see Fig.4 in \cite{donahue2016adversarial}).
In contrast, our GIN is easier to tune, and more importantly, achieves noticeable improved visual performance -- better reconstruction results with no mode dropping even in this small-data situation. 
The reason for this difference contributes to the essentially different objectives of the two methods.
BiGAN uses one discriminator to supervise both the generator and the encoder for representation learning purpose or even latent regression \cite{dumoulin2016adversarially}.
On the contrary, the objective of our GIN is to find the best inverse mapping for efficient sampling. Therefore, sequential order of training $G(\cdot)$ and $E(\cdot)$ is implemented in the proposed GIN to ensure the sample-to-sample inverse is \textit{explicitly} trained by MSE metric.
Therefore, we leave out the comparison of BiGAN for downstream classification.

\subsubsection{AISEL framework}
\label{Sec:Synnative}

\begin{figure}[]
\centering
\includegraphics[width=0.49\textwidth]{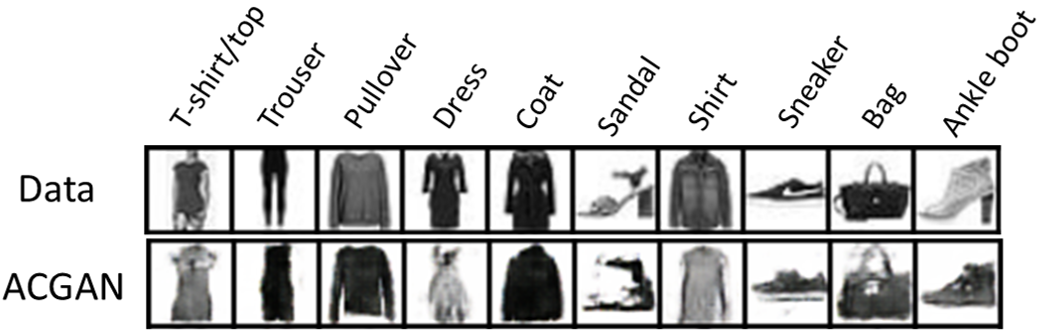}
\caption{\label{fig:SynthACGAN} Qualitative results of generated images of all ten classes via ACGAN baseline.}
\end{figure}

Now we test the rest of our AISEL framework. The native model $C(\cdot)$ is set to be a CNN with detailed architecture specified in Appendix \ref{App:imp}. The classification accuracy of the native model is only $72.8\%$, since only 400 data are used as the training set. 
We then generate an AISEL dataset with size 400. 
Note that the labels are obtained by the oracle model (using all 60,000 training data and WideResNet architecture, denoted as ``Oracle (all)'' in Table \ref{tab:Syn}), mimicking the process of labeling by a domain expert.
An improved classification model $C^*(\cdot)$ can then be obtained by the actual data and AISEL data. 
Final classification accuracy on the same test set is $81.9\%$, an almost $10\%$ increase. 
This improvement shows that the proposed AISEL framework can indeed improve the predictive accuracy in the classification tasks. The reasons are that (i) the additional knowledge, i.e., labeling by the oracle model, is incorporated in the learning process, and (ii) the proposed sampling method ensures that our AISEL dataset explores the feature space. The latter will be discussed in detail below, with comparison to different baseline methods. Table \ref{tab:Syn} shows the final classification performance of the proposed method on both Fashion and MNIST, compared to the baselines.

\textbf{Features of AISEL dataset.}
We first visualize the actual features (i.e., the embedded features of the actual images) in the feature space $\mathbb{F}=[-1,1]^2$. Fig.\ref{fig:SynPredict} (a) shows the 400 actual features (in black crosses) on $\mathbb{F}$, with the background visualizing the uncertainty measure \eqref{equ:entropy}. Specifically, yellow regions indicate high uncertainly of the native model $C(\cdot)$.
Fig.\ref{fig:SynPredict} (b) shows the features for 400 AISEL dataset (in red circles) together with the actual features. 
We observed the key trade-off as mentioned: our AISEL features jointly (i) exploits the highly uncertain regions and (ii) explores the whole feature space. On one hand, objective (i) is achieved by sampling more points in the regions where uncertainty is high, i.e., those with a yellow background. Visually, the AISEL features approximately follow the uncertainty measure. On the other hand, objective (ii) is achieved by spreading the AISEL features over the whole feature space with no features too close to one another. Visually, there are no big ``holes'', and no two points overlap. Therefore, our AISEL method achieves a high ($81.9\%$) classification accuracy.

\textbf{Comparison with GIN-based random sampling.} We compare the proposed sampling method to random sampling. Specifically, we uniformly sample 400 features, generate virtual images using those features, and then label them using the same oracle model (i.e., Oracle (all)).
Fig.\ref{fig:SynPredict} (c) shows the randomly generated features.
We see that those features are (i) not exploiting (i.e., placing more points in) the highly uncertain regions, and (ii) overlapping with one another and to the actual features, leading to poor exploration. Therefore, the classification accuracy of the 400 randomly sampled virtual images is only $76.4\%$, which is noticeably lower ($81.9\%-72.8\%=9.1\%$) than the proposed AISEL method. If increasing the number of random virtual images to 5000, the classification accuracy can be increased to $80.7\%$. Our AISEL method achieves slightly higher accuracy ($81.9\%-80.7\%=1.2\%$), however much less virtual images, and therefore much lower labeling cost.

\textbf{Comparison with active learning.} 
We compare the proposed sampling method to an active learning (AL) method. Specifically, we adapt a similar setting in GAAL \cite{zhu2017generative} in literature, using the GIN to generate a potential unlabeled dataset.
To sample a virtual dataset, we set a grid (with size $101\times 101$) on the feature space, and then select the top 400 features among the grid, whose uncertainty is the highest. 
Virtual images are then generated and labeled by the oracle model.
Fig.\ref{fig:SynPredict} (d) shows the features selected by our setting of AL.
We observe that, though the selected features locate in the highly uncertain regions, they are too close to one another. Furthermore, the selected features do not explore the whole feature space. The final classification performance of this AL is $78.2\%$, which is slightly better than the random sampling ($78.2\%-76.4\%=1.8\%$).
However, our AISEL method, jointly explore and exploit the feature space, achieves a better classification performance ($81.9\%-78.2\%=3.7\%$).

\textbf{Comparison with ACGAN.} Another approach also for small-data tasks is the ACGAN-based data augmentation method \cite{frid2018synthetic}, which generates a set of images based on the chosen labels.
We train an ACGAN \cite{odena2017conditional} using the actual dataset at hand. The complexity of the ACGAN is set to be similar to our GIN. Fig.\ref{fig:SynthACGAN} visualizes the generated images of all ten classes. We see that due to the limited training size (400 in total), the images can sometimes be wrongly labeled -- in Fig.\ref{fig:SynthACGAN}, the generated ``Trouser'' is visually more close to ``Dress''. The final classification accuracy is $72.3\%$, i.e., it offers similar classification accuracy as the native model. 
This is because adapting ACGAN, the labels of the augmented dataset are obtained by the training set \textit{without} complementary information. 
The data size is increased, however, those labels may not be accurate; therefore, little improvement is observed in this experiment.
In our AISEL framework, we use an additional labeling method (i.e., by the oracle model with an accuracy of $96.7\%$) for more precise labels. Therefore, our method achieves better predictive performance.

\textbf{Comparison with transfer learning.} Transfer learning is also popular for small-data tasks. In our setting of transfer learning, we adapt a pre-trained ResNet18 \cite{he2016deep} (by ImageNet \cite{ImageNet_cvpr09}) and only fine-tune the last fully connected layer using the training data (400 in total) at hand. The classification accuracy of the transfer learning is $74.3 \%$, only slightly better than the native model with an accuracy of $72.8\%$. 
The reason for this is that images of the Fashion dataset are virtual different from the natural images in the ImageNet. 
This observation is typical in the applications of manufacturing and healthcare, where the input images are, e.g., images from a thermal camera or flow velocity contour.
In the transfer learning setting \cite{pan2010survey}, we implicitly assume the parameters learned by ImageNet data can use be used to interpret the current Fashion dataset at hand. 
From the result of classification accuracy, the above assumption may not be valid. 
Our AISEL framework incorporates more accurate knowledge from physical experiments or experts, and therefore better classification model can be obtained. 

\textbf{Comparison with native model using 800 actual data.} Another interesting baseline, though not feasible in real applications, is directly using 800 \textit{actual} data to train an oracle model. In our setting, the first 400 data is the same as the 400 for the native model, and the remaining 400 is again randomly selected from the actual training set of the Fashion dataset. The same architecture as the improved model $C^*(\cdot)$ is used. We observe the classification accuracy is $81.3\%$ (denoted as ``oracle (800)'' in Table \ref{tab:Syn}), which is similar to our AISEL method with accuracy $81.9\%$. This is again due to our efficient sampling method, which both explores and exploits the image space. Meanwhile, this also verifies the good generating performance of our GIN.

\subsection{Aortic stenosis application}
We now go back to the motivating application of aortic stenosis (AS). An anonymous image dataset containing 168 patients with aortic stenosis is collected (by Piedmont Healthcare, Atlanta). For each patient, pre-surgical CT scans and the corresponding calcification amount are acquired. The learning task is to classify the calcification level as high or low, which is an important yet challenging clinical problem. Four-fold cross validation strategy is used (see Appendix \ref{App:imp}), leading to only 126 data as the training set.
We first provide more background information on the medical problem and our dataset. We then visualize the GIN performance, with a focus on the pathophysiological meaning. Finally, we discuss the classification accuracy, compared with baselines.

\begin{figure}[]
\centering
\includegraphics[width=0.48\textwidth]{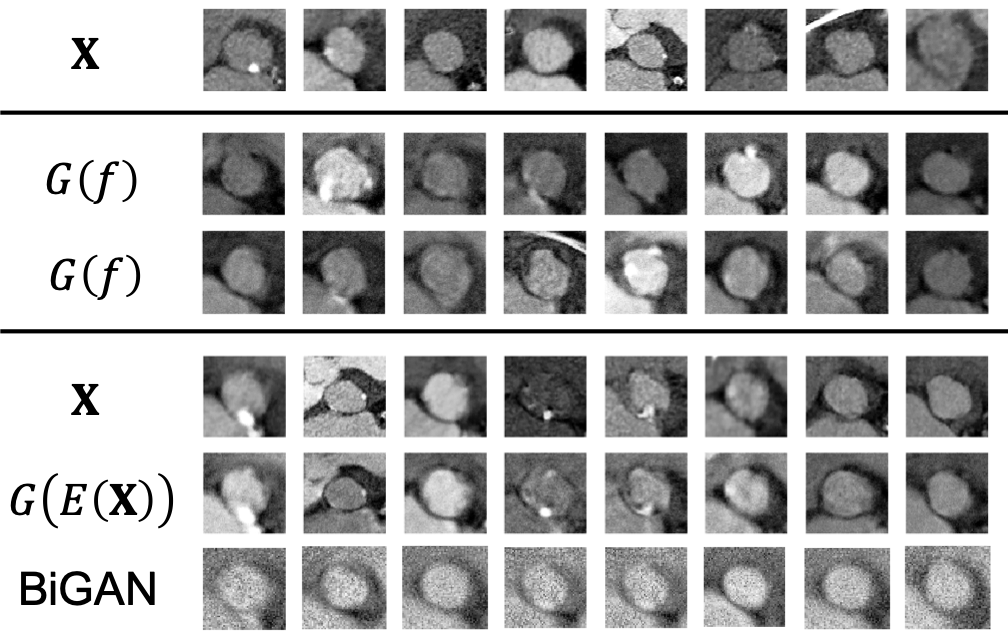}
\caption{\label{fig:Reconstruction} Qualitative GIN results for the aortic stenosis application, including actual data $\bm{X}$, generated samples $G(f)$, and corresponding reconstructions $G(E(\bm{X}))$.}
\end{figure}

\subsubsection{Background on aortic stenosis}
\label{Sec:data}
Aortic stenosis (AS) is one of the most common and most serious valvular heart diseases. Transcatheter aortic valve replacement (TAVR) is a less-invasive treatment option for severe AS patients who are at high risk for open-heart surgery. One of the major post-procedural complications of TAVR is the paravalvular leakage (PVL), i.e., blood flow leakage around the implanted artificial valve due to incomplete sealing between the implant and the native aortic valve, which is often caused by the \textit{calcifications} presented at the aortic annulus region (a ring-shaped anatomic structure connecting the left ventricle and the aortic valve). Therefore, in clinical practice, the amount and the distribution of annular calcifications are of great importance to predicting the occurrence of post-TAVR PVL. 
However, in-vitro study \cite{qian2017quantitative} is quite costly, requiring expensive operation costs of CT scanner, as well as several days of an experimenter's time per virtual patient.
Because of this, we simplify the task of PVL prediction to the task of calcification evaluation, which is deemed as an important clinical indicator of PVL risk. 
Due to the variant contrast level in the aortic root and the fast motion of the valve leaflets, it remains challenging to accurately evaluate calcification near the aortic annulus in pre-TAVR CT images.

\subsubsection{Pathophysiological interpretability of GIN}
\label{sec:exp1}

We first visualize the performance of GIN. 
Here, the dimension of the feature space is $r=20$, i.e., $\mathbb{F}=[-1,1]^{20}$, considering the complexity of the CT scans. 
The detailed architecture of the GIN can be found in Appendix \ref{App:imp}.

\begin{table*}[]
\caption{A comparison of classification accuracy (accu., \%), sensitivity (sens., \%), specificity (spec., \%), and F1 score (\%) of the native model and different improved models in a 4-fold cross-validation, with data size included.}
\label{tab:Accu}
\begin{tabular}{c|cccc|cccc|cccc|cccc}
\toprule
 & \multicolumn{4}{c|}{\textbf{Native Model (126)}} & \multicolumn{4}{c|}{\textbf{Randomly Generated (+1134)}} & \multicolumn{4}{c|}{\textbf{ AISEL (+1134)}} & \multicolumn{4}{c}{\textbf{Randomly Generated (+10000)}} \\ \hline
\textit{{Fold}} & Accu.     & Sens.      & Spec.  & F1    & Accu.       & Sens.      & Spec.     & F1  & Accu.        & Sens.          & Spec.      & F1      & Accu.       & Sens.        & Spec.   & F1    \\ \hline
\textit{{1}}    & 64.29       & 52.17          & 78.95      & 61.54    & 69.05        & 60.87          & 78.95       & 68.29    & 76.19          & 73.91             & 78.95      & 74.42       & 78.57        & 78.26           & 78.95    & 77.27 \\
\textit{{2}}    & 56.96       & 47.26          & 66.67      & 48.65     & 64.29        & 61.90          & 66.67     & 60.00      & 73.81          & 76.19             & 71.43     & 71.43       & 76.19        & 71.43           & 80.95   & 73.17\\
\textit{{3}}    & 57.14       & 50.00          & 63.64      & 52.63     & 71.43        & 52.94          & 84.00      & 58.82     & 80.95          & 76.47             & 84.00       & 76.92      & 85.71        & 82.35           & 88.00      & 82.05     \\
\textit{{4}}    & 59.52       & 55.00          & 63.64    & 56.41         & 64.29        & 60.00          & 68.18   & 60.00        & 71.43          & 75.00             & 68.18       & 69.77     & 73.81        & 70.00           & 77.27    & 70.00   \\ \hline
\textit{{Ave}}  &  \underline{\textbf{59.48}}   & 51.11          & 68.23      & 54.81     &  \textbf{67.27}        & 58.93          & 74.45       & 61.78   &  \underline{\textbf{75.60}}           & 75.39             & 75.64           & 73.14   &  \textbf{78.57}         & 75.51           & 81.29   & 75.62 \\
\toprule
\end{tabular}
\end{table*}

\begin{figure}[!t]
\centering
\includegraphics[width=0.49\textwidth]{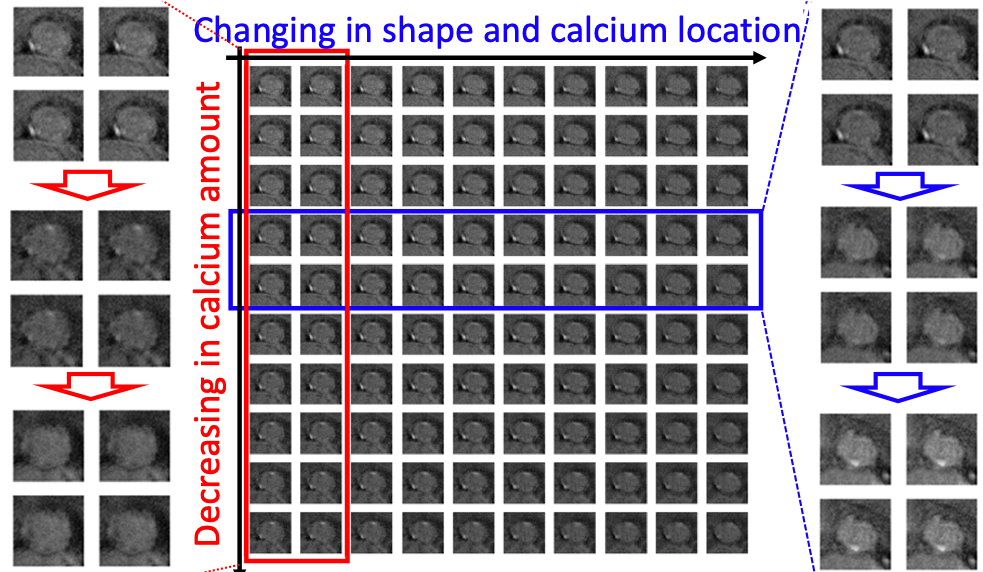}
\caption{\label{fig:FeatureSpace} Qualitative visualization of 2D cross-section of feature space with the generated virtual images on the (partial) grid of feature space. The full and enlarged version of the figure is shown in Fig.\ref{fig:FeatureSpaceAll} in appendix \ref{App:Design}.}
\end{figure}

\textbf{Pathophysiologically-interpretable feature space.}
To better visualize the 20-dimensional feature space $\mathbb{F}=[-1,1]^{20}$,
Fig.\ref{fig:FeatureSpace} shows a randomly selected 2D cross-section of $\mathbb{F}$ with the generated virtual valve images located at their projected feature locations; the full and enlarged version of the figure is shown in Fig.\ref{fig:FeatureSpaceAll} in appendix \ref{App:Design}.
We notice that the variation of the virtual images on the feature grids is continuous and smooth. 
Furthermore, we observe that the two axes of the 2D cross-section shown in Fig.\ref{fig:FeatureSpace} (also see Fig.\ref{fig:FeatureSpaceAll})  have pathophysiological meaning. As shown in the red box (enlarged images on the left side), the vertical axis can be interpreted as the change of the calcification (i.e., the regions of high intensity in the CT images) amount. As shown in the blue box (enlarged images on the  right side), the horizontal axis can be interpreted as the change of valve shape and the calcification location. Similar observations can be found in the other vertical or horizontal groups of images, which demonstrate the potential pathophysiological interpretability of $\mathbb{F}$.

\textbf{Reconstruction test.}
Similar to the toy experiments, Fig.\ref{fig:Reconstruction} shows the reconstruction test comparing the actual CT scans $\bm{X}$ and the reconstructed images $G(E(\bm{X}))$. 
Visually, the reconstructed images are almost identical to the actual images with similar background color and valve geometry. 
Furthermore, the most important pathophysiological indicators, i.e., the location and size of the calcifications are well-recovered. This shows that: (i) using GIN can capture the features of important pathophysiological meaning, and (ii) $G(\cdot)$ and $E(\cdot)$ are inverses of each other. In Fig.\ref{fig:Reconstruction}, we also compare the proposed GIN with BiGAN (see Section 5.1.2), with better performance observed.

\subsubsection{Improving classification by our AISEL method}
\label{sec:exp2}

Here, we use the CT scans at the annulus to predict the calcification level (see Section \ref{Sec:data}). 
The native model utilizes a simple CNN structure, with the detailed architecture described in Appendix \ref{App:imp}. 
Table \ref{tab:Accu} summarizes the classification accuracy, sensitivity and specificity of the four-fold cross validation using the native model. 
For each fold, we generate an AISEL dataset with the size of 1134.
In addition, two randomly sampled virtual dataset, with the size $1134$ and $10000$ are also generated as comparison. 
We will leave out the other baselines, since a detailed comparison is already conducted in the toy experiments (see Section 5.1).
As for labeling the virtual patients, an empirical approach is performed: 
a mixture model of two Gaussians is used to model the pixel intensity, based on whether the pixels are classified as normal tissues or calcifications. 
The volume of the calcification region is then calculated.
After that, a manual check is performed by a radiologist and the calcification levels are corrected if needed.
Note that if budget allows, a more sophisticated labeling approach can be used.

\textbf{Classification performance.}
The three generated virtual datasets (proposed, random with size 1134 and 10000) are fused with the actual dataset to obtain improved classifiers. 
Table \ref{tab:Accu} summarizes the prediction accuracy together with the sensitivity, specificity, and F1 score of the different classifiers. 
We see that the native model performs the poorest over the test set, with less than $60\%$ averaged accuracy. 
The prediction accuracy improves to $67.27\%$ when using randomly generated samples with the size of 1134. 
Using our AISEL method, the averaged accuracy improves to $75.60\%$ with the same size (126 actual + 1134 virtual), a improvement of $15\%$ against the native model and $8\%$ compared to the random sampling method. 
Furthermore, if increasing the size of the randomly generated virtual dataset to 10000, which may lead to overly expensive labeling costs, the prediction accuracy is higher, but not noticeably higher, than our AISEL dataset with a size of $1260$. 
As a summary, promising results in Table \ref{tab:Accu} suggest: (i) the proposed AISEL method efficiently incorporate physical knowledge, and therefore yields better prediction performance; (ii) with the same data size, the proposed sampling strategy, both exploring and exploring the design space, leads to a better downstream classifier than random sampling; and (iii) small AISEL dataset can achieve similar predictive performance compared to a much bigger randomly generated dataset, which reduces the labeling cost of conducting physical experiments.

\section{Conclusion and future work}
\label{sec:Discussion}
In this paper, we proposed the AISEL framework to efficiently sample a virtual dataset to incorporate complementary \textit{physical knowledge} for small-data learning problems, with applications to manufacturing and healthcare. 
We first propose a novel generative invertible network (GIN), which can find the bidirectional mapping of generating virtual images and extracting the features of the actual images. 
We then propose a new sampling strategy, which both explores and exploits the image space to minimize the predictive uncertainty.
Our AISEL method can achieve better performance in toy experiments, compared to the state-of-the-art baselines.
Furthermore, in the motivating applications of aortic stenosis, our method lowers the labeling cost by $90\%$ while achieving a $15\%$ improvement in prediction accuracy.

Looking ahead, we are pursuing several directions for future research.
From a methodological point of view, we are interested in other approaches to incorporating physical knowledge. Methods in \cite{chen2019function,raissi2017physics1} appear to be attractive options.
In the application point of view, further study of predicting post-surgical blood pattern is of interest.
While our method can still be used, the difficulties lie in the physical labeling process.
Tissue-mimicking 3D printing technology \cite{chen2018efficient} and in-vitro studies \cite{qian2017quantitative} appear to be suitable.

\ifCLASSOPTIONcompsoc
  \section*{Acknowledgments}
\else
  \section*{Acknowledgment}
\fi

The authors would like to thank Dr. Shizhen Liu at Piedmont Heart Institute in Atlanta, GA, for providing medical knowledge. The authors also want to thank Dr. Zih Huei Wang at Feng Chia University in Taiwan and Dr. Geet Lahoti at Kabbage, Inc. for the insightful discussions.

\ifCLASSOPTIONcaptionsoff
  \newpage
\fi

\bibliographystyle{IEEEtran}

\onecolumn
\newpage

\appendices

\numberwithin{equation}{section}


\section{Proof of Theorem 1.}
\label{App:thm1}

Since the generator $G(\cdot)$ is obtained by (\ref{equ:trainGIN}) with the training error $<\epsilon$, i.e., 
\begin{equation}
\mathcal{W}(\mathcal{X}, G_\#[\mathcal{U}]) = \inf_{\gamma} \int _{\mathbb{X}\times \mathbb{X}}  \|x-G(u)\|_2 d \gamma(x,G(u)) =d<\epsilon.
\end{equation}
This means we have obtained the transportation map $\gamma: \mathbb{X} \times \mathbb{X} \mapsto[0,1]$, s.t.,
\begin{equation}
\mathbb{E}_\gamma[\|X-G(U)\|_2] = \int  \|x-G(u)\|_2 d \gamma(x,G(u)) =d,
\end{equation}
For any realization $x_i\in \mathbb{X}$ of the random variable $X \sim \mathcal{X}$, one can find a $u_i \in \mathbb{F}$ using the following optimization scheme:
\begin{equation}
u_i =\argmin_{u\in \mathbb{F}}\|x_i-G(u)\|_2,
\label{equ:findU}
\end{equation}
We denote this as $u_i=h(x_i)$. If we denote the conditional measure of $\gamma$ as $\gamma_{x_i}=\gamma|X=x_i$. Given $X=x_i$ and $u_i=h(x_i)$, clearly, we have, 
\begin{equation}
\|x_i-G(u_i)\|_2 \leq \mathbb{E}_{U\sim \gamma_{x_i}}\|x_i-G(U)\|_2,
\label{equ:LesCond}
\end{equation}
Furthermore, recall the dual formula of the Wasserstein distance:
\begin{equation}
\mathcal{W}(\mathcal{X}, G_\#[\mathcal{U}])= \sup_{\|D(\cdot)\|_L\leq 1} \mathbb{E}_{x\sim \mathcal{X}}[D(x)] - \mathbb{E}_{u\sim \mathcal{U}}[D(G(u))]<\epsilon,
\end{equation}
Specifically, if let the function $D(x)= h(x)-E(x)$, we have :
\begin{equation}
 \|\mathbb{E}_{x\sim \mathcal{X}}[h(x)-E(G(h(x)))] - \mathbb{E}_{{u}\sim {\mathcal{U}}}[u-E(G(u))]\|<\epsilon,
 \label{equ:DuW}
\end{equation}
With $x_i$, $u_i$ and the Lipschitz-L continues assumption on $G(\cdot)$, we have:
\begin{equation}
   \|G(E(x_i))-x_i\|_2 \leq \|G(E(x_i))-G(u_i)\|_2+\|G(u_i)-x_i\|_2 \leq L\|E(x_i)-u_i\|_2+\|G(u_i)-x_i\|_2
\end{equation}
Now, replace the realization $x_i$ with the random variable $X$ and take the expectation over $X \sim \mathcal{X}$,
\begin{equation}
   \mathbb{E}_{X\sim \mathcal{X}}\|G(E(X))-X\|_2 \leq L\mathbb{E}_{X\sim \mathcal{X}}\|E(X)-h(X)\|_2+\mathbb{E}_{X\sim \mathcal{X}}\|G(h(X))-X\|_2.
\end{equation}
Considering the way we choose $u_i=h(X_i)$ and the inequality (\ref{equ:LesCond}), for the second term above, we have
\begin{equation}
   \mathbb{E}_{X\sim \mathcal{X}}\|X-G(h(X))\|_2 \leq \mathbb{E}_{X\sim \mathcal{X}}\mathbb{E}_{U\sim \gamma_X}\|X-G(U)\|_2 = \mathbb{E}_{\gamma}\|X-G(U)\|_2 \leq \epsilon
\end{equation}
As for the first term, we have:
\begin{equation}
   \mathbb{E}_{X\sim \mathcal{X}}\|E(X)-h(X)\|_2 \leq \mathbb{E}_{X\sim \mathcal{X}}\|E(X)-E(G(U))\|_2+\mathbb{E}_{X\sim \mathcal{X}}\|h(X)-E(G(U))\|_2
\end{equation}
With the Lipschitz-L continues assumption on $E(\cdot)$:
\begin{equation}
   \mathbb{E}_{X\sim \mathcal{X}}\|E(X)-h(X)\|_2 \leq L\mathbb{E}_{X\sim \mathcal{X}}\|X-G(U)\|_2+\mathbb{E}_{X\sim \mathcal{X}}\|h(X)-E(G(U))\|_2\leq L\epsilon+\mathbb{E}_{X\sim \mathcal{X}}\|h(X)-E(G(U))\|_2
\end{equation}
Note that $E(\cdot)$ is obtained by (\ref{Equ:TrainCNN}) with training error $\mathbb{E}_{U\sim \mathcal{U}} \left[\|E(G(U))-U\|_2\right] < \delta $. Recall Equation (\ref{equ:DuW}),
\begin{equation}
\mathbb{E}_{X\sim \mathcal{X}}\|U-E(G(U))\|_2 \leq \mathbb{E}_{U\sim \mathcal{U}}\|U-E(G(U))\|_2 +\epsilon \leq \delta +\epsilon
\end{equation}
Finally,  we have 
\begin{equation}
   \mathbb{E}_{X\sim \mathcal{X}}\|G(E(X))-X\|_2 \leq L(L\epsilon+\delta+\epsilon) +\epsilon = (L^2+L+1)\epsilon+L\delta.
\end{equation}

\section{Proof of Theorem 2.}
\label{App:thm2}

we denote the target measure as $\mathcal{X}$ with its empirical representation as $\mathcal{X}_n$, while $\mathcal{X}'$ as measure obtained by proposed approach with its empirical representation as $\mathcal{X}'_m$. 

(i) As the training data size approach infinity, $\mathcal{X}'_n \to \mathcal{X}_n$, this because using Pinsker’s inequality,
\begin{equation}
\bigg|\sum_{x_i \in \mathcal{D}} \mathcal{I}(x_i>y) -\sum_{x'_i \in \mathcal{D}'} \mathcal{I}(x'_i>y) \bigg|<\sqrt{KL(\mathcal{X}'_n||\mathcal{X}_n)}.
\end{equation}
where $KL(\cdot||\cdot)$ is the K-L divergence of two distribution. From \cite{arjovsky2017wasserstein}, we know the $KL(\mathcal{X}'_n||\mathcal{X}_n)\to 0$ as $n\to \infty$, i.e., existing a small $\epsilon>0$, for any $y$,
\begin{equation}
\bigg|\sum_{x_i \in \mathcal{D}} \mathcal{I}(x_i>y) -\sum_{x'_i \in \mathcal{D}'} \mathcal{I}(x'_i>y) \bigg|<\frac{\epsilon}{3}.
\end{equation}
For more discussion and justification, please refer to \cite{arjovsky2017wasserstein}.

(ii) As the data size approach infinity, $\mathcal{X}_n \to \mathcal{X}$. Since the 
training dataset $\mathcal{D}=\{x_i\}_{i=1}^n$ is sampled from the target measure $\mathcal{X}$, its empirical cumulative distribution function (CDF) converges to target CDF $F_\mathcal{X}$, i.e.,  for any $y$,
\begin{equation}
  \bigg|\sum_{x_i \in \mathcal{D}} \mathcal{I}(x_i>y) - F_\mathcal{X}(y)\bigg|<\frac{\epsilon}{3}.
\end{equation}

(iii) Similar to (ii), as the data size approach infinity, $\big|\sum_{x'_i \in \mathcal{D}'} \mathcal{I}(x'_i>y) - F_\mathcal{x'}(y)\big|<\epsilon/3$.

We have the difference in the obtained CDF and the target CDF:
\begin{equation}
  \big|F_\mathcal{X}(y)-F_\mathcal{x'}(y)\big|=\bigg|F_\mathcal{X}(y)-\sum_{x_i \in \mathcal{D}} \mathcal{I}(x_i>y) + \sum_{x_i \in \mathcal{D}} \mathcal{I}(x_i>y)-\sum_{x'_i \in \mathcal{D}'} \mathcal{I}(x'_i>y) + \sum_{x'_i \in \mathcal{D}'} \mathcal{I}(x'_i>y) -F_\mathcal{x'}(y)\bigg|.
\end{equation}
Combining (i), (ii) and (iii), we know as the training data size large enough, with any $y$,
\begin{equation}
  \big|F_\mathcal{X}(y)-F_\mathcal{x'}(y)\big| \leq \bigg|F_\mathcal{X}(y)-\sum_{x_i \in \mathcal{D}} \mathcal{I}(x_i>y)\bigg|+\bigg|\sum_{x_i \in \mathcal{D}} \mathcal{I}(x_i>y) -\sum_{x'_i \in \mathcal{D}'} \mathcal{I}(x'_i>y) \bigg|+\bigg|\sum_{x'_i \in \mathcal{D}'} \mathcal{I}(x'_i>y) -F_\mathcal{x'}(y)\bigg|<\epsilon.
\end{equation}
i.e., as the training data size approach infinity, $\mathcal{X}' \to \mathcal{X}$ in distribution.

\section{Proof of Theorem 3.}
\label{App:thm4}
Note that the objective function in \eqref{equ:seq_sp} is not a energy distance. However, we have 
\begin{align}
 &\argmin_{ \{f'_1, \cdots, f'_m\}} \sum_{i=1}^{n} \mathbb{E}_{\gamma \sim \mu_h}\|f'_i-\gamma\|_2 -\frac{1}{2(m+n)}\sum_{i=1}^{m+n}\sum_{j=1}^{m+n} \|f'_i-f'_j\|_2 \\
 =& \argmin_{\{f'_1, \cdots, f'_m\}} \sum_{i=1}^{m+n} \mathbb{E}_{\gamma \sim \mu_h}\|f'_i-\gamma\|_2
 +\sum_{i=m+1}^{m+n} \mathbb{E}_{\gamma \sim \mu_h}\|f'_i-\gamma \|_2 -\frac{1}{2(m+n)}\sum_{i=1}^{m+n}\sum_{j=1}^{m+n} \|f'_i-f'_j\|_2  \\
 =& \argmin_{\{f'_1, \cdots, f'_m\}} dist(\mathcal{F}'_{n+m},\mu_h)
\end{align}
Here, $m$ is the number of virtual points and $n$ is the number of actual points. Here, we set $f'_i=f_i$ for actual points with indies $i= m+1, \cdots, m+n$ and let $\mathcal{F}'_{n+m}$ be the empirical measure for features $\{f'_i\}_{i=1}^{n+m}$. Note that the minimizing is only on the virtual dataset, i.e., with subscripts $1, \cdots, m$.  

Note that for energy distance 
\begin{equation}
    dist(\mathcal{F}'_{n+m},\mu_h)\leq dist(\mathcal{F}'_{n+m},\mathcal{F}'_m)+
    dist(\mathcal{F}'_m,\mu_h)
    \label{eqapp:1}
\end{equation}
As $m\rightarrow \infty$, the second term in \eqref{eqapp:1} approaches to zero, and equivalently we are minimizing $dist(\mathcal{F}'_m,\mu_h)$. Following \cite{mak2018support}, we have  $\mathcal{F}'_m\rightarrow \mu_h$ in distribution.

Furthermore, with the continuity condition on the $G(\cdot)$, we have
\begin{equation}
\mathcal{X}'_m \rightarrow \mu_{H}
\end{equation}
directly following the continuous mapping theorem.

\section{Details of the implementation}\label{App:imp}
We explain the implementation details here.

\textbf{Four-fold cross validation}.
A four-fold cross validation strategy is used in the aortic stenosis application. 
We have 168 data in total. 
Three quarters of the data ($168 \times 75\% \times 10  = 1260$) after rotation augmentation is used as the training set, while the remaining quarter ($168 \times 25\% = 42$) will be the testing set.
Since the architecture of the classifier is pre-defined, and there are no hyperparameters that need to be tuned, the validation set is not needed.

\textbf{Training GIN}.
For the synthetic dataset, the generator $G(\cdot)$ adapts 5-layer vanilla NN with 512, 512, 1024, 1024, 1024 hidden nodes in each hidden layer, respectively, and ReLu activation. 
The discriminator $D(\cdot)$ also adapts 5-layer vanilla NN with 1024, 1024, 1024, 512, 512 hidden nodes, and ReLu activation.
As for the encoder $E(\cdot)$, it has 10 convolutional layers with 128, 256, 256, 512, 512, 1024, 1024, 1024, 512, 256 hidden nodes in each hidden layer, respectively, leaky ReLu activation and batch normalization. In our implementation, we train the GIN for 2000 epochs, with a constant learning rate of $1e-5$.
For the aortic stenosis applications, the architecture and the training strategy are similar, except that the numbers of the hidden nodes and training epochs are doubled.

\begin{figure}[]
\centering
\includegraphics[width=0.9\textwidth]{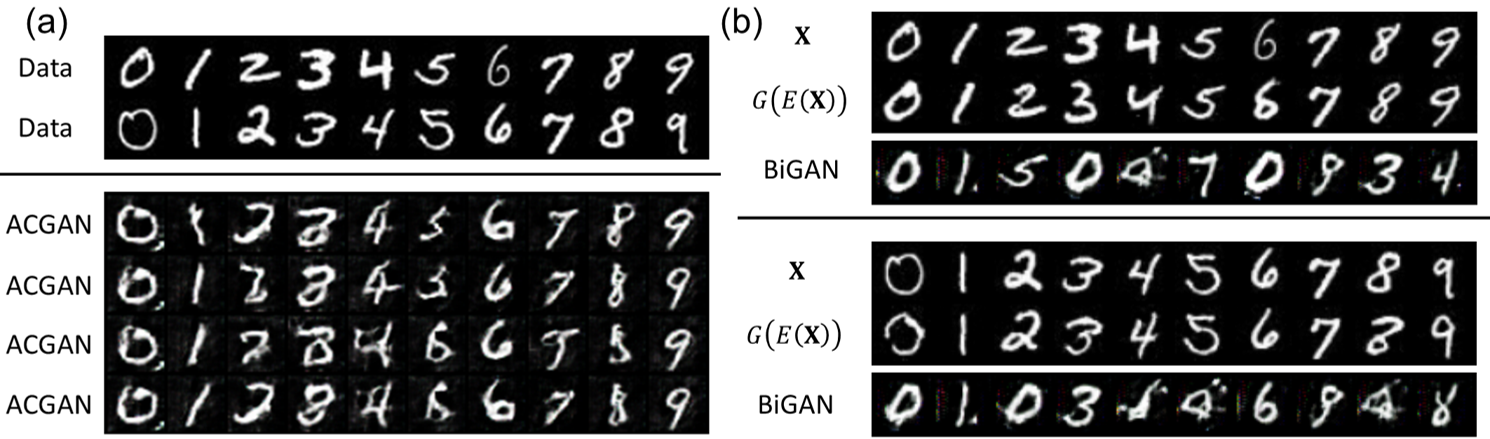}
\caption{\label{fig:SynthRe2} Qualitative results for GIN training using MNIST, including the training set data $\bm{X}$ of different classes, generated samples via ACGAN, our reconstructions $G(E(\bm{X}))$ and reconstructions via BiGAN. }
\end{figure}

\begin{figure}[]
\centering
\includegraphics[width=0.98\textwidth]{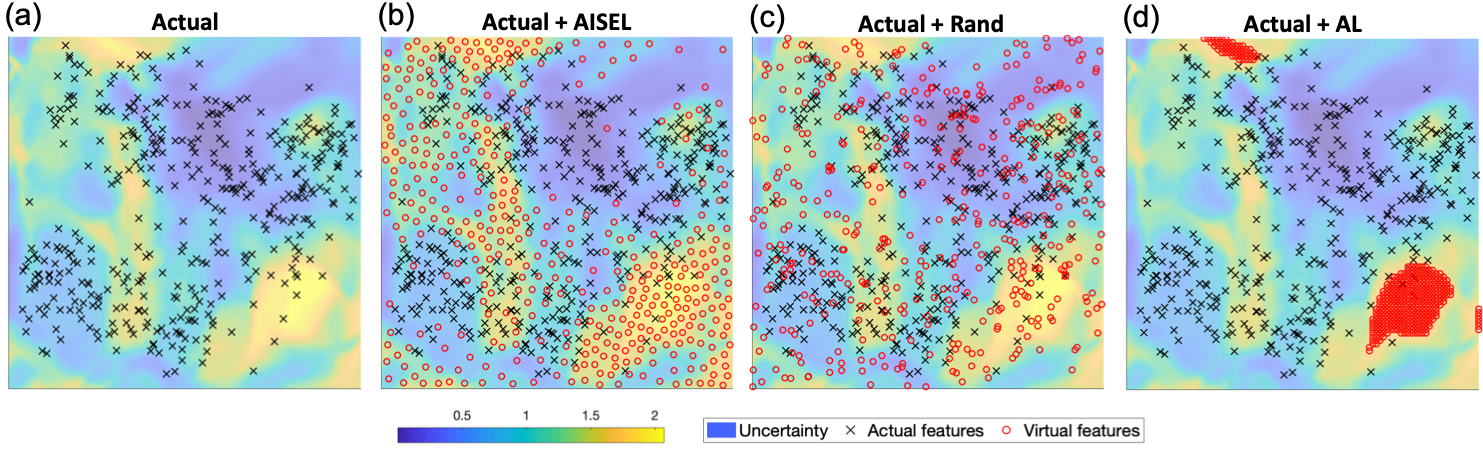}
\caption{\label{fig:SynPredict2} A comparison of the selected features by our AISEL method, the random sampling method, and the active learning method on the MNIST dataset, with the uncertainty measure \eqref{equ:entropy} as background.} 
\end{figure}

\textbf{Training native and improved models}.
For the toy computer vision datasets, both the native model $C(\cdot)$ and the improved model $C^*(\cdot)$ have three convolutional layers with 32, 64 and 64 hidden nodes, respectively. Leaky ReLu activation and batch normalization are also included in each layer. After the convolutional layers, two fully connected layers with 512 and 64 hidden nodes are used, respectively, with ReLu activation and batch normalization. Cross-entropy loss is used. In our implementation, we train the CNN for 80 epochs. The initial learning rate is $1e-4$, with decay to a half every 20 epochs. We select the above for the best empirical performance in preliminary experiments.
For the aortic stenosis application, both $C(\cdot)$ and $C^*(\cdot)$ have the similar three convolutional layers with leaky ReLu activation and batch normalization. The three convolutional layers have 32, 64 and 128 hidden nodes, respectively. After the convolutional layers, three fully connected layers with 512, 128 and 32 hidden nodes are used, respectively, with ReLu activation and batch normalization. Cross-entropy loss and the same decay of learning rate is used. 
Note that the complexity for both $C(\cdot)$ and $C^*(\cdot)$ is low, especially compared to the encoder $E(\cdot)$. This is mainly because of the difference in the classification task for $C(\cdot)$ and regression task for $E(\cdot)$.

\section{Toy MNIST experiments}

\label{app:MNIST}

We conduct the same experiments on the MNIST dataset as the Fashion dataset in Section 5.1; the setup and the implementation details are the same as that in the Fashion experiments.
Fig.\ref{fig:SynthRe2} shows the visual comparison of the proposed GIN and baselines.
We see that in Fig.\ref{fig:SynthRe2} (b), our GIN can generate shape images, with visual superior reconstruction performance than the BiGAN. As for ACGAN (see Fig.\ref{fig:SynthRe2} (a)), we observe that the performance is not as good as the proposed GIN.

The final classification performance is already shown in Table \ref {tab:Syn}. 
Our AISEL method achieves predictive accuracy of $91.2\%$, a $91.2\%-88.2\%=3\%$ improvement compared to the native model. Meanwhile, our method outperforms the baselines, e.g., transfer learning, ACGAN-based method, and active learning. As for the GIN-based random augmentation, our method achieves (i) better performance when the same amount of virtual data (400) is used and (ii) similar performance when 5000 data is used in the baseline. This superior performance is again contributing to the exploration and exploitation of the feature space (see Fig.\ref{fig:SynPredict2}).

\section{Balancing the label distribution}
\label{app:balance}
The proposed sampling method can also be used to balance the label distribution. 
Note that the uncertainty measure defined in \eqref{equ:entropy} is not normalized.  
In order to balance the data, we modify the uncertainty measure as 
\begin{align}
    h_b(f_0)=\frac {h(f_0)}{ \int h(f)  \mathcal{I}\left[c(f) = c(f_0)\right]\; df_0},
 \label{equ:bmu}
\end{align}
where notation $c(f)=\argmax C(G(f))$ denotes the label (rather than the predictive probability) of the native model, $\mathcal{I}[\cdot]$ is the indicator function. The denominator normalizes the density with respect to different classes, and therefore balances the label distribution. 
\begin{table*}[]
\centering
\caption{A comparison of F1 score, area under the receiver operating characteristic curve (AUC), and classification accuracy of the native model and different improved models, under the imbalanced training dataset. 
}
\label{tab:Syn}
\begin{tabular}{c|ccc|ccc}
 & \textbf{Native (300)} &\textbf{Undersampling}  & \textbf{Oversampling} & \textbf{Random (+600)}  & \textbf{AISEL (+300)} & \textbf{AISEL (+600)}   \\ \hline
\textit{F1 score}           & \textbf{0.9448}   & 0.9589    & 0.9624  & 0.9728   & \textbf{0.9734}     & {\textbf{0.9801}} \\ 
\textit{AUC}           & \textbf{0.9782}   & 0.9839    & 0.9846   & 0.9905   & \textbf{0.9937}     & {\textbf{0.9964}} \\ 
\textit{Accuracy}           & {\textbf{94.60\%}}   & 95.90\%     & 96.25\%   & 96.95\%     & \textbf{97.05\%}     & {\textbf{98.00\%}} 
\end{tabular}
\end{table*}

\begin{figure*}[]
\centering
\includegraphics[width=0.98\textwidth]{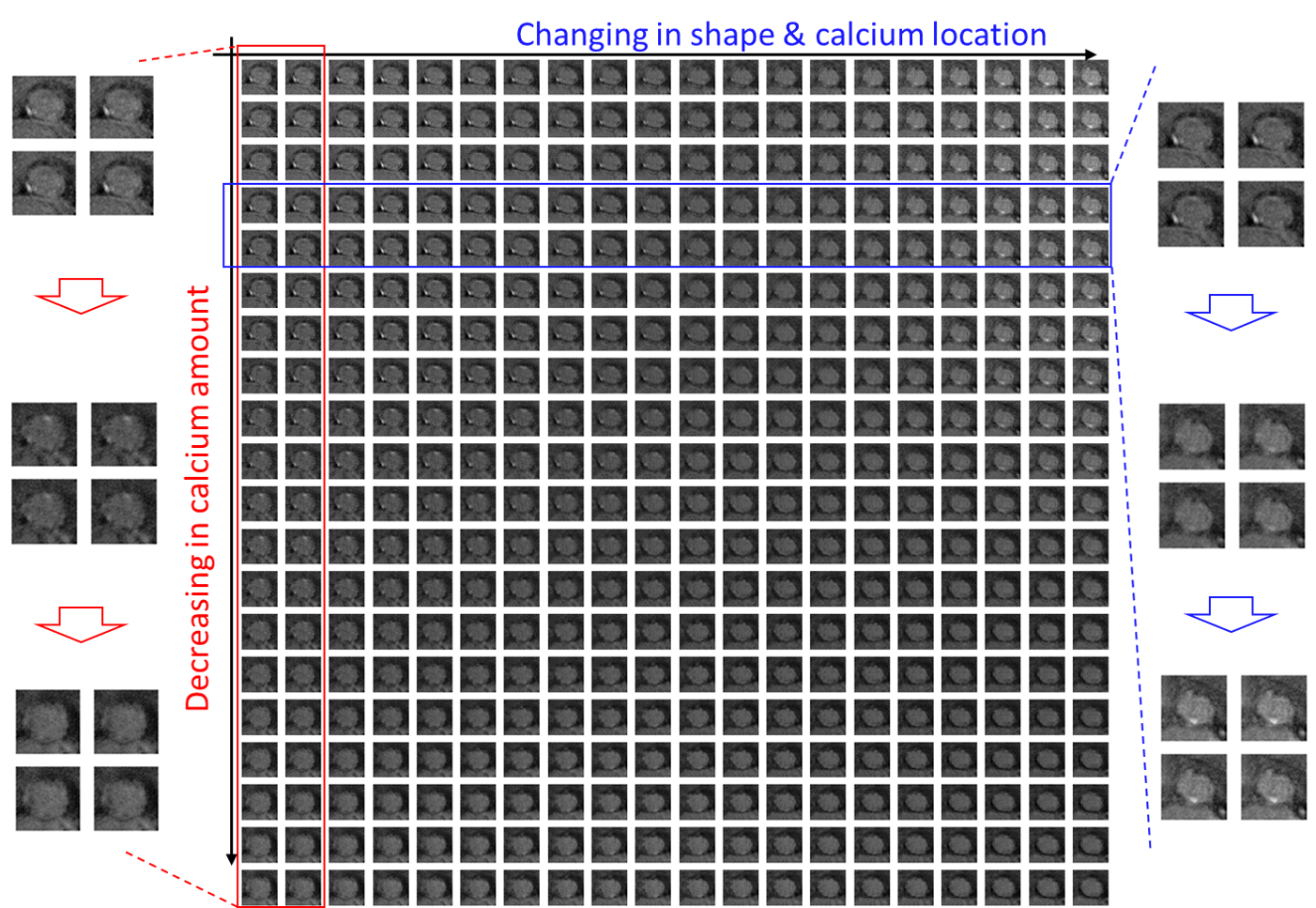}
\caption{\label{fig:FeatureSpaceAll} Qualitative visualization of 2D cross-section of feature space with the generated virtual images on the grid of feature space. The pathophysiological meaning of both axes is visualized in the left and right sides, respectively.}
\end{figure*}

We then apply this to learn a classification model from the truncated and imbalanced Fashion dataset. For the sake of simplicity, we consider two-class classification (i.e., using the two classes ``Top" and ``Coat"). The training dataset is designed to be both small ($n=300$) and imbalanced (size of classes, $11:1$). We then applied the proposed AISEL framework as discussed in Section 5.1.1 using the same setup as discussed in Appendix \ref{App:imp}.  Table \ref{tab:Syn} lists the performance of the proposed AISEL method (with balancing via \eqref{equ:bmu}), random sampling (without balancing), and the standard baselines of under-sampling and over-sampling. Compared with the native model, the two sampling strategies improve predictive performance marginally. In contrast, the improvement in predictive accuracy using the proposed AISEL method is around $2.5\%$ and $3.5\%$ with $300$ and $600$ virtual data points, respectively. Meanwhile, the proposed AISEL method mitigates the imbalance of the training dataset with improvements of at least $0.028$ and $0.015$ in the F1 score and AUC, respectively. 
Furthermore, compared to the randomly generated virtual dataset of size $600$, the proposed AISEL method (with balancing) achieves noticeable improvements, even with a smaller data size of $300$.

\section{More on aortic stenosis application}\label{App:Design}

Due to the limited space, Fig.\ref{fig:FeatureSpace} in Section 5.2.2 only shows the partial 2D cross-section of the feature space. Fig.\ref{fig:FeatureSpaceAll} visualizes the whole and enlarged 2D cross-section feature space. The two axes of the 2D cross-section shown in Fig.\ref{fig:FeatureSpaceAll} have pathophysiological meaning. As shown in the red box (enlarged images on the left side), the vertical axis can be interpreted as the change of the calcification (i.e., the regions of high intensity in the CT images) amount. As shown in the blue box (enlarged images on the  right side), the horizontal axis can be interpreted as the change of valve shape and the calcification location.

\begin{figure*}[]
\centering
\includegraphics[width=0.98\textwidth]{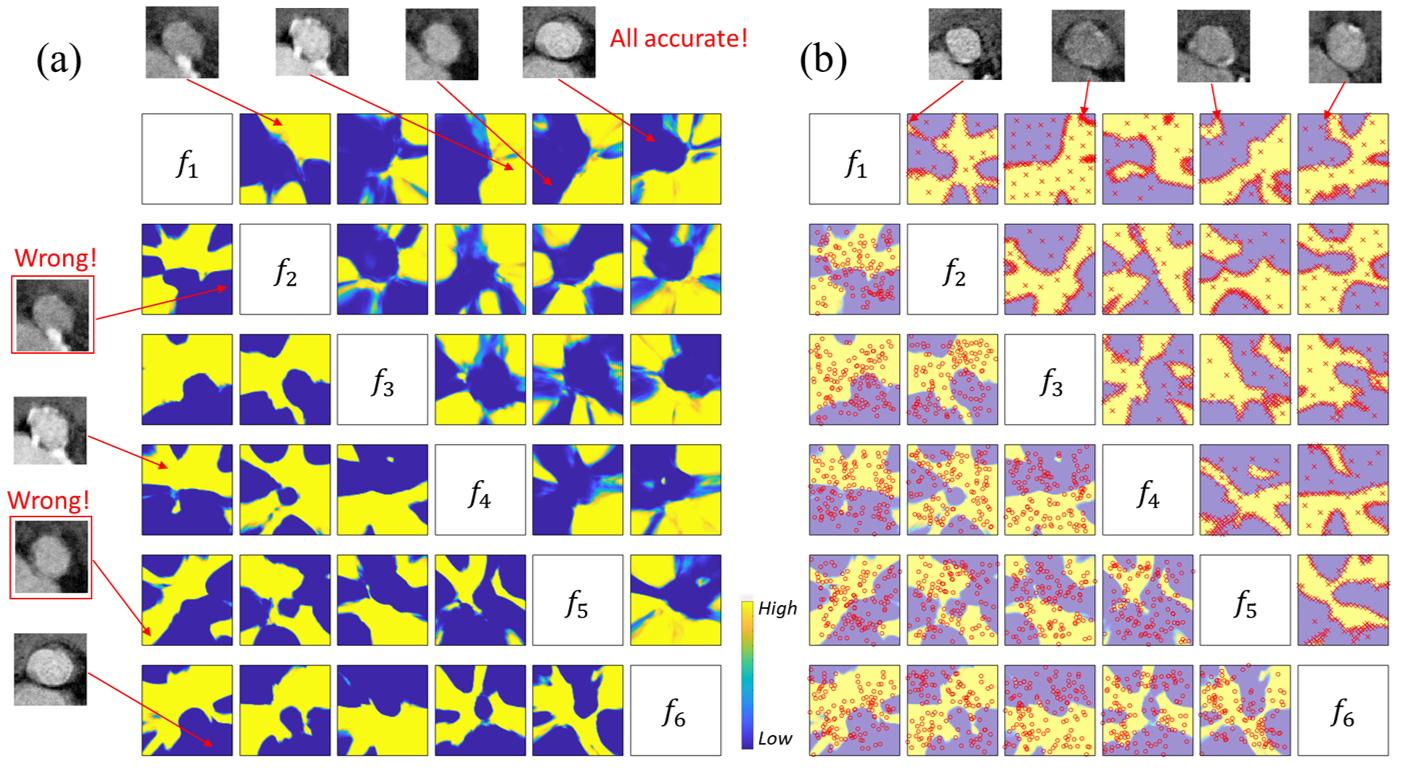}
\caption{\label{fig:PredictCal} (a) A compression of the classification model of the native model (bottom left) and the improved model (top right) via proposed method on the 6D feature space $\mathbb{F}$. Four testing images are show and the native model can only correctly classify two of them, while the improved model can correctly classify all of them. (b) A compression of the actual patients (bottom left) and the selected features of our AISEL dataset (top right) in the feature space. Four examples of the generated virtual patients are also shown.}
\end{figure*}

In order to better visualize the sampled AISEL dataset and demonstrate how it helps in improving the classification accuracy, we conduct the experiments in the Section \ref{sec:exp2} with the dimension of the feature space $r=6$, i.e., $\mathbb{F}=[-1,1]^6$.
Furthermore in the two-class classification problem, we use $C(G(\cdot)):\mathbb{F} \mapsto [0,1]$, with low value $\sim 0$ denoting the low calcification situation and high value $\sim 1$ for high calcification situation. 
The prediction contour $C(G(\cdot))$ of the model learned by the first three folds of the training data (the remaining fold is for testing) is shown in the lower-left half of Fig.\ref{fig:PredictCal} (a). Every small figure visualizes a 2D subspace of $\mathbb{F}$ with the remaining features set to be zero. Note that all combinations of the two features (totally 15 for six features) are shown in the lower-left half of Fig.\ref{fig:PredictCal} (a). Yellow means high calcification (i.e., $C(G(f))=1$), while blue means low calcification (i.e., $C(G(f))=0$).
Meanwhile, four testing valves are shown in the left side of Fig.\ref{fig:PredictCal} (a). The obtained native model $C(\cdot)$ only accurately classifies two of them, which indicates the poor performance of the native model.

Designed AISEL data are shown in upper left region of Fig.\ref{fig:PredictCal} (b). Note that for every figure, only $10\%$ of the  feature of AISEL dataset closest to the 2D cross-section plane are included for better visualization purpose. Most of AISEL features, as expected, are located on the boundary of the prediction contour of the native model, while the rest features are uniformly spread over the whole space. This again shows the exploration and exportation properties of the proposed AISEL method.
Four examples of the selected virtual images in the AISEL dataset are also visualized on the top, with an arrow pointing their features in the feature space.
Visually, they are indeed confusing for predicting the calcification amount. After physical labeling by a radiologist, they will help improve the classifier.
As a comparison, the actual images (totally, 126) projected in every 2D cross-section are shown in the lower right region of Fig.\ref{fig:PredictCal} (b). Note that the actual images are randomly distributed in the whole 6D space with no apparent pattern. 

The prediction contour $C^*(G(\cdot))$ of the improved classifier $C^*(\cdot)$ using our AISEL method is shown in the top left half of Fig.\ref{fig:PredictCal} (a). We can see the finer structure is learned indicating a more sophisticated model is obtained. 
Meanwhile, the four characteristic images tested by the native model is also tested by the $C^*(\cdot)$. The classification of all four is accurate, showing a noticeable improvement in the prediction accuracy.

\end{document}